\documentclass[letterpaper]{article} 
\usepackage{aaai25}  
\usepackage{times}  
\usepackage{helvet}  
\usepackage{courier}  
\usepackage[hyphens]{url}  
\usepackage{graphicx} 
\usepackage{natbib}  
\usepackage{caption} 
\usepackage{subfigure}
\usepackage{svg}
\usepackage{amsmath}
\usepackage{pifont}
\usepackage{color}
\usepackage{multirow}
\usepackage{booktabs}
\usepackage{xcolor}
\usepackage{colortbl}
\usepackage{soul}
\usepackage{floatrow}
\usepackage{cleveref}

\newfloatcommand{capbtabbox}{table}[][\FBwidth]
\usepackage{algorithm}
\usepackage{algorithmic}

\usepackage{newfloat}
\usepackage{listings}
\DeclareCaptionStyle{ruled}{labelfont=normalfont,labelsep=colon,strut=off} 
\floatstyle{ruled}
\newfloat{listing}{tb}{lst}{}
\floatname{listing}{Listing}
\title{Reverse Region-to-Entity Annotation for Pixel-Level Visual Entity Linking}
\author {
    Zhengfei Xu\textsuperscript{\rm 1},
    Sijia Zhao\textsuperscript{\rm 1},
    Yanchao Hao\textsuperscript{\rm 2},
    Xiaolong Liu\textsuperscript{\rm 2},\\
    Lili Li\textsuperscript{\rm 2},
    Yuyang Yin\textsuperscript{\rm 2},
    Bo Li\textsuperscript{\rm 2},
    Xi Chen\textsuperscript{\rm 2},
    Xin Xin\textsuperscript{\rm 1}\footnote{Xin Xin (xxin@bit.edu.cn) is the corresponding author.}
}
\affiliations {
    \textsuperscript{\rm 1}School of Computer Science and Technology, Beijing Institute of Technology, Beijing, China\\
    \textsuperscript{\rm 2}Platform and Content Group, Tencent, Beijing, China\\
    \{zhengfei, sijiazhao, xxin\}@bit.edu.cn, 
    \{marshao, loongliu, irinali, yuyangyin, ryanbli, jasonxchen\}@tencent.com
}

\begin{document}

\maketitle

\begin{abstract}
Visual Entity Linking (VEL) is a crucial task for achieving fine-grained visual understanding, matching objects within images (visual mentions) to entities in a knowledge base. Previous VEL tasks rely on textual inputs, but writing queries for complex scenes can be challenging. Visual inputs like clicks or bounding boxes offer a more convenient alternative. Therefore, we propose a new task, Pixel-Level Visual Entity Linking (PL-VEL), which uses pixel masks from visual inputs to refer to objects, supplementing reference methods for VEL. To facilitate research on this task, we have constructed the Mask\textsc{Oven}-Wiki dataset through an entirely automatic reverse region-entity annotation framework. This dataset contains over 5 million annotations aligning pixel-level regions with entity-level labels, which will advance visual understanding towards fine-grained. Moreover, as pixel masks correspond to semantic regions in an image, we enhance previous patch-interacted attention with region-interacted attention by a visual semantic tokenization approach. Manual evaluation results indicate that the reverse annotation framework achieved a 94.8\% annotation success rate. Experimental results show that models trained on this dataset improved accuracy by 18 points compared to zero-shot models. Additionally, the semantic tokenization method achieved a 5-point accuracy improvement over the trained baseline. 
\end{abstract}

\begin{links}
    \link{Datasets}{https://github.com/NP-NET-research/PL-VEL}
\end{links}

\section{Introduction}
Visual Entity Linking (VEL) is an open-domain visual entity recognition task that expands the label space to web-scale knowledge bases. As a key task for achieving fine-grained visual understanding, VEL contributes to various tasks such as multimodal knowledge graphs completion \cite{wuRecognizingUnseenObjects2023}, visual question answering (VQA) \cite{qiuSnapNTellEnhancingEntityCentric2024a}, image caption \cite{zhangEndToEndSpatiallyConstrainedMultiPerspective2024}, image retrieval \cite{sainCLIPAllThings2023,saitoPic2WordMappingPictures2023} and so on.

\begin{figure}[ht]
    \centering
    \includegraphics[width=0.86\columnwidth]{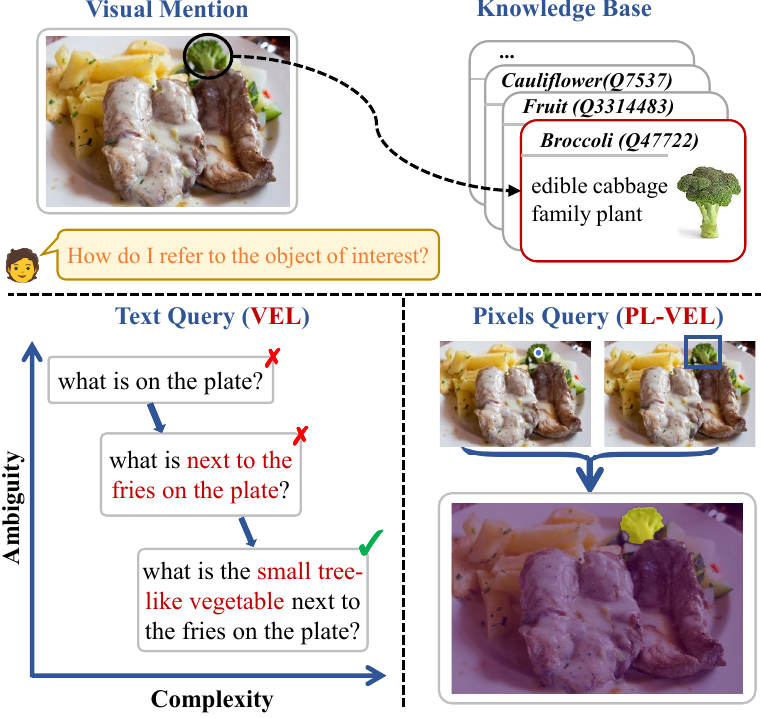} 
    \caption{Overview of comparing text and pixel-based Visual Entity Linking (VEL) tasks}
    \label{fig:task_overview}
\end{figure}

Current VEL tasks \cite{huOpendomainVisualEntity2023,caronGenerativeApproachWikipediaScale2024,xiaoGroundingLanguageModels2024a} relying on textual queries struggle with some complex scenes. For example, in \cref{fig:task_overview}, a simple query like \textit{what is on the plate?} cannot accurately refer to \texttt{Broccoli}, requiring more complex queries, such as \textit{what is the small tree-like vegetable next to the fries on the plate?} Creating such queries demands extensive background knowledge and precise comprehension of visual relationships. This adds an additional burden on users, and we cannot assume that downstream models are equipped with such capabilities.

\begin{figure*}
\centering
    \subfigure[]{
        \includegraphics[height=0.236\textwidth]{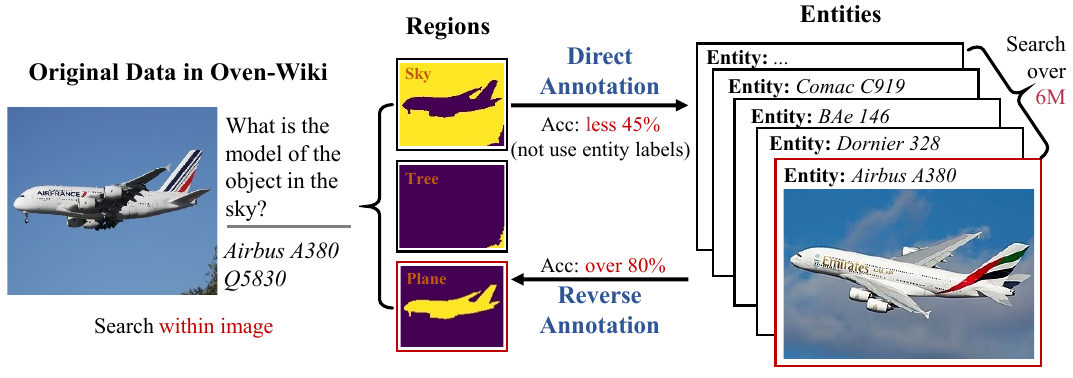}
        \label{fig:reverse_motivation}
    }
    \subfigure[]{
        \includegraphics[height=0.236\textwidth]{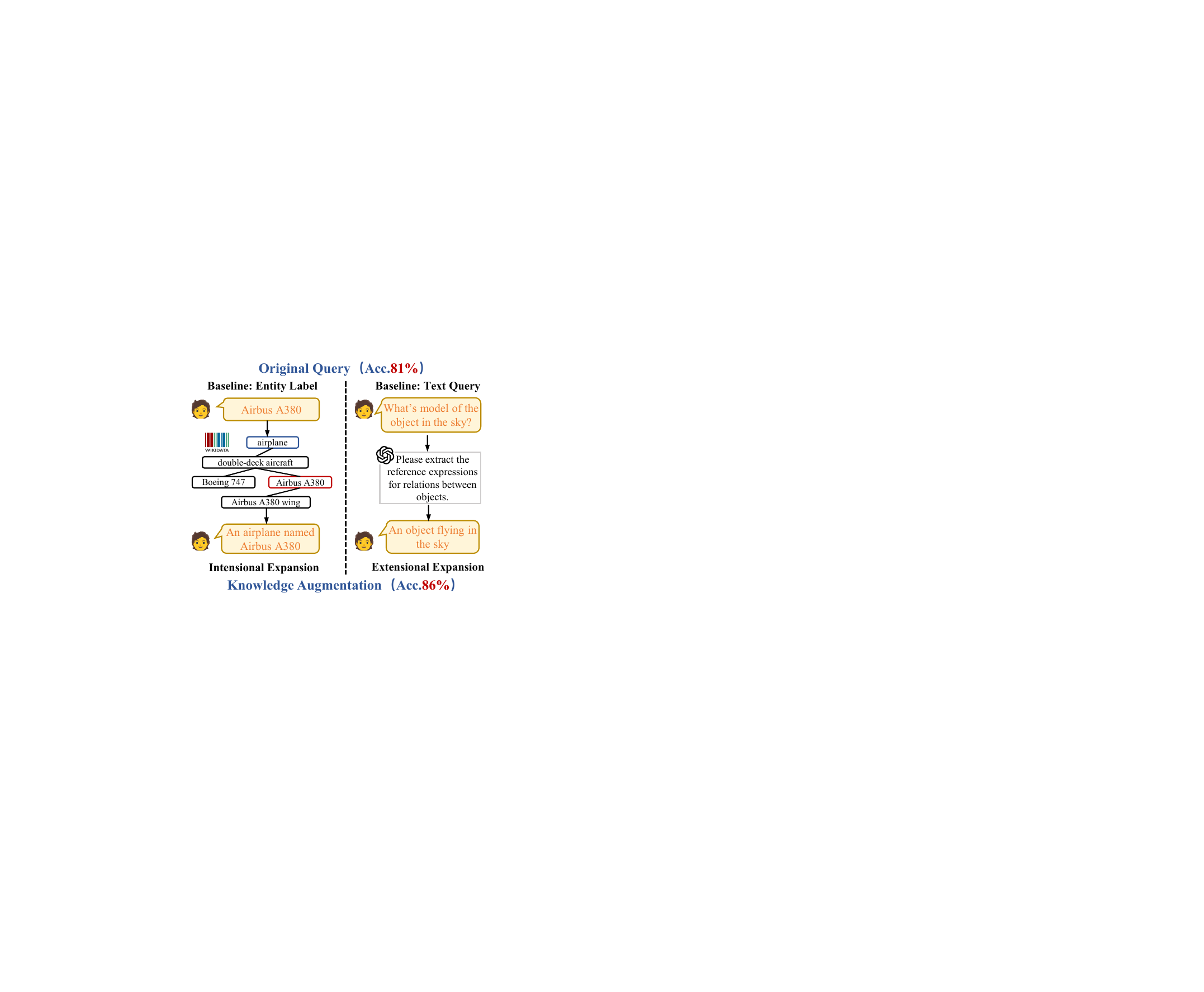}
        \label{fig:ka_motivation}
    }
\caption{Overview of the annotation framework. (a) Comparison of direct and reverse annotation shows that direct annotation struggles to utilize existing entity labels effectively, whereas reverse annotation efficiently reduces the search space. (b) Knowledge-enhanced text prompt for segmentation models, built on intensional and extensional expansion.}
\label{fig:anno}
\end{figure*}

In such complex scenes, visual prompts such as clicks, boxes, and pixel masks can be supplementary methods for more efficient and accurate reference. Therefore, this work introduces Pixel-Level Visual Entity Linking (PL-VEL), which uses pixel masks to refer to visual mentions and link them to knowledge-base entities, as shown in \cref{fig:task_overview}. With promptable segmentation models like SAM \cite{kirillovSegmentAnything2023} and SEEM \cite{zouSegmentEverythingEverywhere2023}, users or downstream models can easily create pixel masks through simple actions such as clicking, and drawing boxes. It makes PL-VEL more practical than traditional VEL tasks in real-world applications, such as VQA\cite{qiuSnapNTellEnhancingEntityCentric2024a} and visual reasoning \cite{Chen_2024_CVPR}. To support the research on this task, a large-scale open-domain PL-VEL dataset that aligns pixel-level mask regions in images with entities in a knowledge base is required. 

The straightforward approach to constructing the dataset follows the VEL setup, \textbf{mapping visual objects to entities} by segmenting everything in the images and mapping each region to its corresponding entity. However, using GPT-4V to generate the entity names and searching within 6M entities achieves only about 25\% accuracy \cite{xiaoGroundingLanguageModels2024a}. Even powerful text-based VEL model \textsc{Auto}VER-13B \cite{xiaoGroundingLanguageModels2024a} reaches only around 45\% accuracy, making constructing a high-quality dataset challenging.

To address these challenges, we adopt a reverse approach by \textbf{mapping entities to visual objects}, as illustrated in \cref{fig:reverse_motivation}. We constructed the Mask\textsc{Oven}-Wiki dataset based on the existing \textsc{Oven}-Wiki dataset \cite{huOpendomainVisualEntity2023}, which aligns entities with images. By segmenting pixel regions based on entity labels, we provide pixel references for visual mentions. This reverse annotation method leverages existing labels and reduces the search space from millions of entities to image regions. Pre-experiments with the segmentation pipeline model Grounded-SAM \cite{renGroundedSAMAssembling2024a} show an annotation accuracy of approximately 80\%. 

Despite this, understanding long-tail entities remains challenging for segmentation models. We introduce a two-part knowledge augmentation method to improve annotation quality, as shown in \cref{fig:ka_motivation}. For \textbf{intensional expansion}, we retrieve hypernyms from Wikidata\footnote{https://www.wikidata.org} to provide broader semantics for entities in queries. For \textbf{extensional expansion}, we use GPT-3.5 to extract referring expressions from the original text questions that contain spatial or semantic relationships. This augmentation improves annotation accuracy from 81\% to 86\%. To address error propagation in the segmentation pipeline, we adopt the end-to-end model SEEM \cite{zouSegmentEverythingEverywhere2023}. We employ a model ensemble and heuristic rules to filter and correct low-quality annotations, thereby achieving an accuracy of approximately 95\%. Finally, we developed a PL-VEL dataset with 5M visual mentions.

The PL-VEL task is more challenging than existing VEL tasks because it does not rely on textual queries with strong prior. To enhance visual feature utilization and region-interacted attention, we propose a visual semantic tokenization method based on Osprey \cite{yuanOspreyPixelUnderstanding2023}. Our approach produces more independent and complete image tokens than the fixed-size image patch sequence in ViT \cite{dosovitskiyImageWorth16x162020}. Experiments show our method improves model accuracy by about 5 points.

In summary, our main contributions are as follows:   
\begin{itemize}
    \item We introduce the PL-VEL task and construct Mask\textsc{Oven}-Wiki, a large-scale dataset aligning pixel-level regions with entity-level labels.
    \item We design a reverse annotation framework that achieves 94.8\% annotation accuracy through knowledge augmentation and model ensemble.
    \item We establish a PL-VEL baseline, achieving an accuracy improvement from 1.3\% to 25.2\% by fine-tuning on Mask\textsc{Oven}-Wiki.
\end{itemize}

\section{Related Work}
\paragraph{Visual Entity Linking.} Previous studies, such as Tag2Text \cite{huang2024tagtext} and RAM \cite{zhang2024recognize}, generated common category tags for images but failed to recognize entity-level tags. To address this, \textsc{Oven}-Wiki \cite{huOpendomainVisualEntity2023} was proposed as an open-domain visual entity linking benchmark, which links regions of interest to 6M Wikipedia\footnote{https://www.wikipedia.org/} entities based on text queries. This benchmark also validated the effectiveness of the generative entity recognition framework (GER). Building on this, GER-ALD \cite{caron2024generative} demonstrated that unAmbiguous Language-based Discriminative (ALD) entity codes offer a performance advantage within the GER framework. \textsc{Auto}VER \cite{xiaoGroundingLanguageModels2024a} achieved an accuracy 11.9 points higher than GER-ALD on the \textsc{Oven}-Wiki test set through retrieval-augmented constrained decoding.

In contrast to text-based references, Wikiperson \cite{sunVisualNamedEntity2022}, a VEL dataset using bounding box references, was introduced. However, Wikiperson is limited to ``person" entities and is limited in scale. To address this, we propose an open-domain PL-VEL task, for advancing fine-grained visual understanding.

\paragraph{Region-specific Visual Understanding.} It focuses on semantic information in local image regions, including region-specific conversation \cite{Rasheed_2024_CVPR}, region captioning \cite{yuanOspreyPixelUnderstanding2023}, and referring expressions comprehension \cite{guo2024regiongpt}. Our PL-VEL is also a region-specific recognition task. 
Recent works on region-specific visual understanding focus on MLLMs. Although MLLMs like BLIP \cite{liBLIPBootstrappingLanguageImage2022}, LLaVA \cite{NEURIPS2023_6dcf277e}, and MiniGPT-4 \cite{zhuMiniGPT4EnhancingVisionLanguage2023a} extend LLMs' capabilities to vision. However, they struggle to comprehend effectively specific visual regions. Kosmos-2 \cite{pengKosmos2GroundingMultimodal2023a} and Shikra \cite{chenShikraUnleashingMultimodal2023} input bounding boxes as location-aware reference tokens into LLMs, while GPT4RoI \cite{zhangGPT4RoIInstructionTuning2024} and GlaMM \cite{Rasheed_2024_CVPR} use specialized visual modules for bounding box regions. 

These models, however, cannot describe pixel-level features accurately. Osprey \cite{yuanOspreyPixelUnderstanding2023} achieves pixel-level understanding with a mask-aware visual extractor. Expanding on this, we introduce cross-attention interactions of pixel-level features and train the model on Mask\textsc{Oven}-Wiki to enhance pixel-level visual understanding and provide a baseline for PL-VEL.

\section{Pixel-Level Visual Entity Linking Task}

\subsection{Task Definition}
\paragraph{Original Task (PL-VEL)}
\textit{The PL-VEL task takes an image $I$ and a pixel mask $m$ as input. The pixel mask $m$ represents a visual object in $I$, referred to as a visual mention $V^m$. The goal of PL-VEL is to link this visual mention $V^m$ to its corresponding entity $e$ in the knowledge base $\mathcal{K}$. }

\paragraph{Reverse Annotation (Dataset Construction)}
\textit{The dataset construction task is the reverse process of the PL-VEL task. Given an entity $e$, an image $I$ containing $e$, and a text query $q$ for $e$, it takes them as input, and its goal is to segment the pixel mask $m$ of the visual object of the entity $e$ in $I$.}

The PL-VEL task assumes that mask references for visual mentions are provided. Various visual and textual prompts can be processed into pixel masks using preprocessing models such as SAM \cite{kirillovSegmentAnything2023} and SEEM \cite{zouSegmentEverythingEverywhere2023}. This integration enhances the PL-VEL system's adaptability and supports interactive and fine-grained visual entity comprehension.

\subsection{The Mask\textsc{Oven}-Wiki Dataset Construction}
\label{sec:data_construct}
To define and address the PL-VEL task, we have developed the Mask\textsc{Oven}-Wiki dataset, a benchmark with approximately 5 million annotations, covering various categories of entities. Each annotation includes an image, a visual mention represented by a pixel mask, a text query, and the corresponding entity label from Wikipedia.

For the source of data, we use an open-domain entity recognition dataset, \textsc{Oven}-Wiki \cite{huOpendomainVisualEntity2023}, where each sample includes an image, a text query for visual mention and its corresponding entity. This dataset uses a 6 million-entity set derived from Wikipedia. The dataset aggregates 14 existing datasets and is divided into two subsets based on the original tasks of the source datasets. \textbf{entity split} (ES) for image recognition/retrieval and \textbf{query split} (QS) for visual question answering. Additionally, \textsc{Oven}-Wiki provides a high-quality evaluation dataset, the \textbf{human set}, which is manually annotated. Based on this data, we developed and employed an automated method to annotate pixel-mask visual references for visual mentions in those three subsets. Additionally, we enriched it by annotating visual mentions for entities with images on Wikipedia pages. This additional content serves as a supplement to the knowledge base, referred to as \textbf{wiki split} (WS).

\begin{figure}[ht]
    \centering
    \includegraphics[width=\columnwidth]{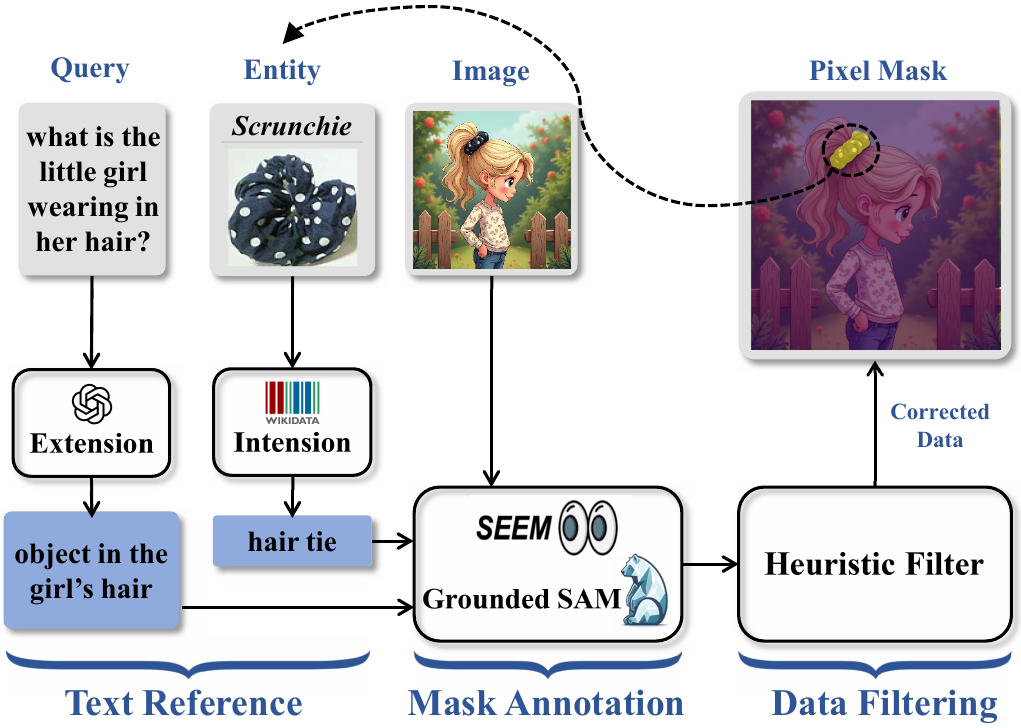}
    \caption{The procedure of building Mask\textsc{Oven}-Wiki. The illustration image is generated by AI \cite{chang2024fluxfastsoftwarebasedcommunication}.}
    \label{fig:construst}
\end{figure}

As illustrated in \cref{fig:construst}, we have developed a knowledge-enhanced methodology for segmentation annotation. This workflow consists of three steps: text reference construction, mask annotation, and data filtering. For automated pixel-mask annotation, we utilize Grounded-SAM \cite{renGroundedSAMAssembling2024a} and SEEM \cite{zouSegmentEverythingEverywhere2023}, which generate pixel masks from textual references. Finally, we apply a heuristic rule-based filter to remove or correct noisy data.

\paragraph{Text Reference Construction.} We construct text references for each visual mention to guide the annotation model. A straightforward approach is to directly input the entity label and text query into the segmentation annotator, but this approach has limitations. Specifically, long-tail entities challenge the annotator's generalization performance. Therefore, we propose a two-part knowledge augmentation method to enhance the text reference.

For intensional description, we enrich the intensional description of entity labels by querying Wikidata. Specifically, we retrieve super-categories of the entity using two properties: \texttt{Instance of (P31)} and \texttt{Subclass of (P279)}. These super-categories are then combined with the original entity information through predefined templates to generate intension-enhanced textual references.

For extensional relations, we leverage relationships between objects to resolve ambiguities. Such relationships are often encoded in the text queries in \textsc{Oven}-Wiki \cite{huOpendomainVisualEntity2023}, for example, \texttt{What is the brown item on the chair facing the camera?}. We use GPT-3.5 to analyze these queries and extract expressions describing the mention's spatial or relational context. This process generates extension-enhanced references, such as \texttt{the brown item on the chair facing the camera}.

\paragraph{Mask Annotation.} We utilize two open vocabulary segmentation models, Grounded-SAM \cite{renGroundedSAMAssembling2024a} and SEEM \cite{zouSegmentEverythingEverywhere2023}, for annotating masks with textual references. Grounded-SAM, as a pipeline tool, initially employs Grounding-DINO \cite{liu2023grounding} to identify bounding boxes based on text prompt, followed by the SAM \cite{kirillovSegmentAnything2023} for segmentation. This pipeline achieves a labeling success rate of 81.4\% in preliminary experiments, forming the foundation of our solution. On the other hand, SEEM, as an end-to-end model, is good at processing diverse inputs. We utilize it as a complementary strategy to mitigate potential error propagation in Grounded-SAM's annotation process.

\paragraph{Data Filtering.} Upon analyzing the results, we have identified four primary issues: reference to non-visual entities, error propagation in the segmentation pipeline, incomplete entity depiction in images, and foreground-background confusion in dense object scenes. To improve the annotation quality, we have applied heuristic filtering rules, as follows: 
\begin{itemize}
    \item For non-visual entities, we deleted the annotations of specific entities, such as events, technology, games, chart reasoning, and so on.
    \item For error propagation in the pipeline, we identify and correct potential errors by analyzing the agreement between different types of reference and segmentation models using intersection over union (IOU) metrics. IOU values indicate potential errors, we correct these by sampling the most confident bounding box using the intersection with segmentation results.
    \item For incomplete entity depiction, we found that such errors mainly occur in location entities. To address this, we apply a confidence threshold constraint specifically for location entities and treat the entire image as the corrected mask.
    \item For foreground-background confusion, we found that such errors mainly occur in dense object scenes. To mitigate this, we employ a rule-based correction using morphological operations. When multiple bounding boxes of the same type cover a significant portion of the image, we apply erosion and dilation to the predicted mask. We then analyze the number of connected components to judge this error and invert the mask for correction.
\end{itemize}

\subsection{The Mask\textsc{Oven}-Wiki Dataset Analysis}
\paragraph{Annotation Quality.} To evaluate the efficacy of our annotation method, we randomly sampled 2,000 annotations for manual inspection. To ensure diversity, we limited 
each entity to a maximum of one sample and proportionally allocated samples from the entity, query, and wiki splits. As shown in \cref{tab:anno_evaluation}, our knowledge-enhanced text references and model-ensemble heuristic filtering rules improved the annotation accuracy from 81\% to 95\%.

\begin{table}[ht]
\centering
    \begin{tabular}{clcc}
    \toprule
    \multicolumn{2}{c}{\multirow{2}{*}{\textbf{Reference}}} & \multicolumn{2}{c}{\textbf{Model}} \\ \cmidrule(r){3-4}
    \multicolumn{2}{c}{} &  G-SAM & SEEM \\ \midrule
    \multirow{2}{*}{original query} & \# entity label & 81.4 & 69.3 \\
     & \# text query & 71.3 & 62.8 \\ \midrule
    \multirow{2}{*}{\begin{tabular}[c]{@{}c@{}}knowledge \\ augmentation\end{tabular}} & \# intension & 86.1 & 65.5 \\
     & \# extension & 83.0 & 67.5 \\ \midrule
    \multicolumn{2}{c}{overall after filtering} & \multicolumn{2}{c}{94.8} \\ \bottomrule
    \end{tabular}
\caption{Annotation accuracy under different settings.}
\label{tab:anno_evaluation}
\end{table}

\begin{figure}[ht]
    \centering
    \includegraphics[width=0.7\columnwidth]{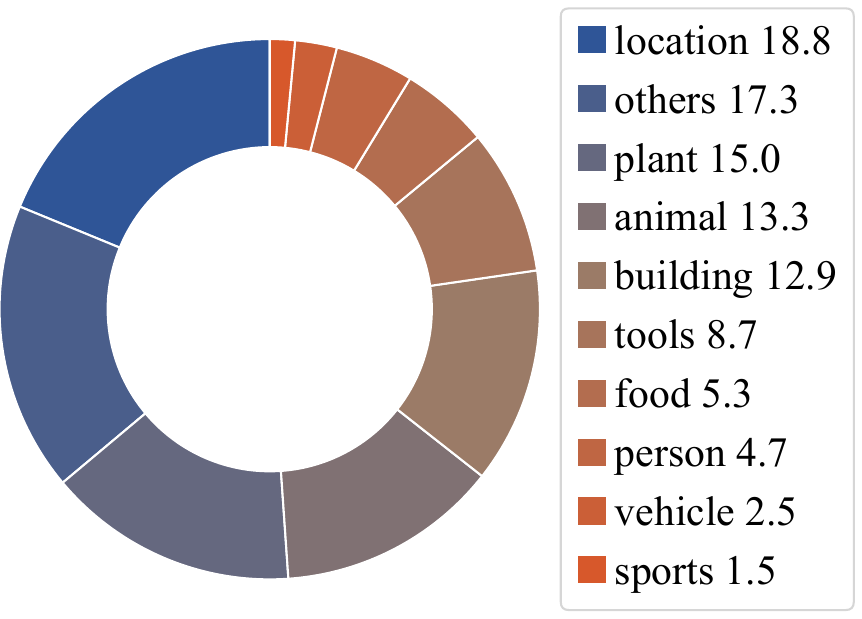}
    \caption{Entity category distribution in the evaluation set.}
    \label{fig:category_distribution_manual}
\end{figure}

\begin{figure*}
\centering
    \subfigure[]{
        \includegraphics[height=0.275\textwidth]{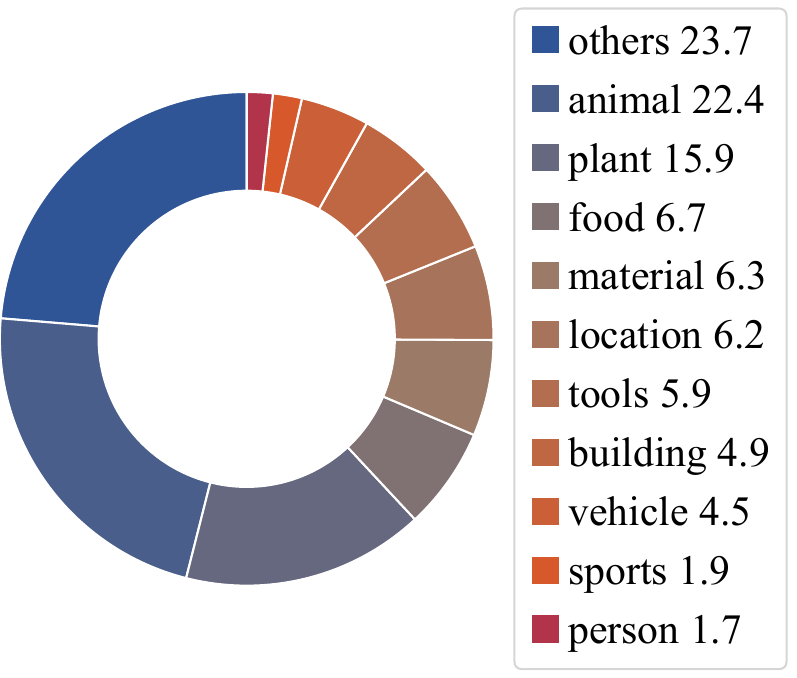}
        \label{fig:distribution_a}
    }
    \subfigure[]{
        \includegraphics[height=0.275\textwidth]{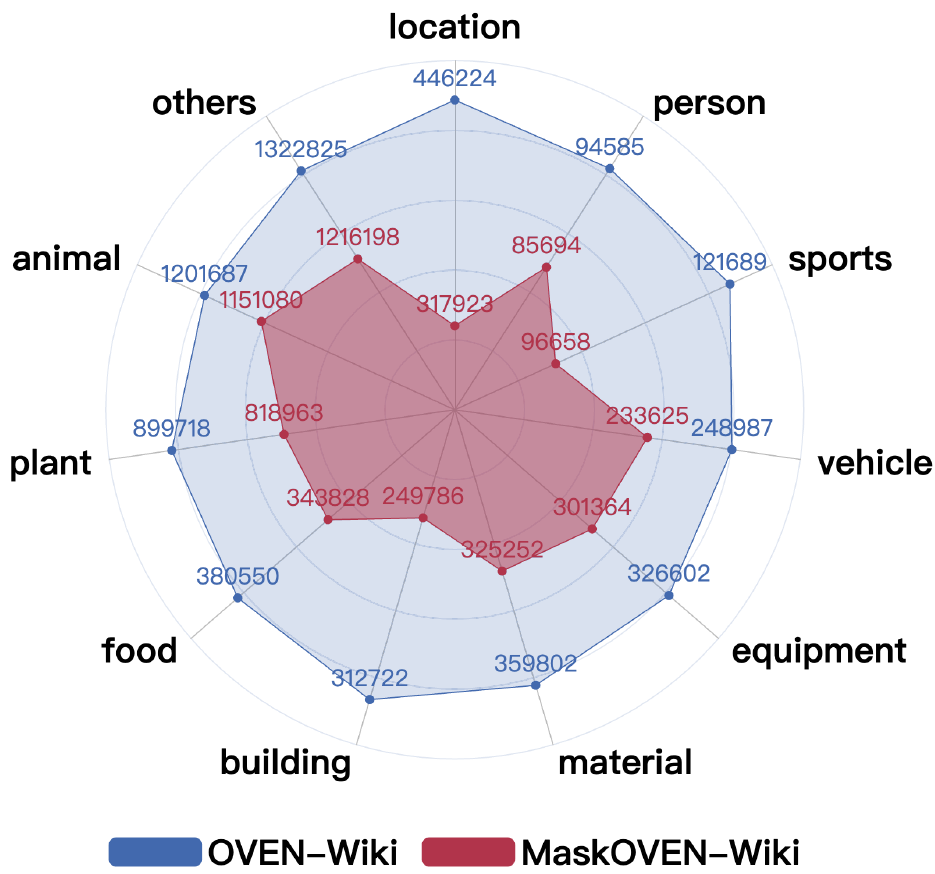}
        \label{fig:comparison}
    }
    \subfigure[]{
        \includegraphics[height=0.275\textwidth]{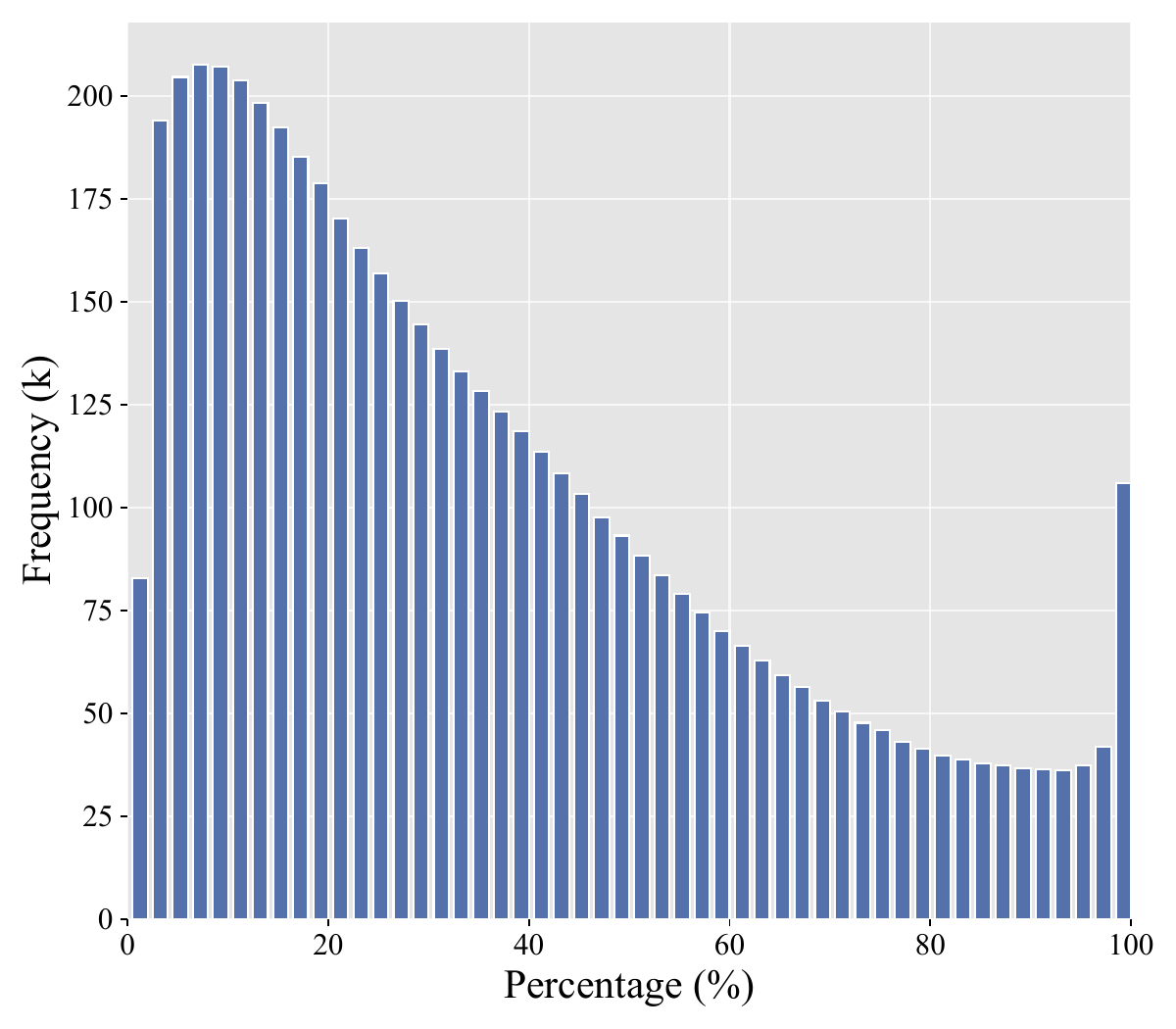}
        \label{fig:ratios}
    }
\caption{Distribution of Mask\textsc{Oven}-Wiki: (a) distribution of entity categories; (b) comparison of the entity category distribution between Mask\textsc{Oven}-Wiki and \textsc{Oven}-Wiki; (c) distribution of mask ratios for visual mentions in images.}
\label{fig:comparison and distribution}
\end{figure*}

Figure \ref{fig:category_distribution_manual} shows the distribution of entity types in this sample set. Compared to \cref{fig:distribution_a}, the entity category distribution of the sample set is similar but more balanced.

Table \ref{tab:overview} summarizes the statistics of Mask\textsc{Oven}-Wiki dataset. Our dataset contains 5,245,421 annotations for 5,214,965 images from \textsc{Oven}-Wiki \cite{huOpendomainVisualEntity2023} dataset covering 20,077 entities. We reused the knowledge base of \textsc{Oven}-Wiki, which contains 6,063,945 Wikipedia entities, of which 2,032,340 entities have a corresponding image.

\begin{table*}
    \centering
    \begin{tabular}{lrrrrrrrr}
        \toprule
         & \multicolumn{2}{c}{\textbf{Train Set}} & \multicolumn{2}{c}{\textbf{Val Set}} & \multicolumn{2}{c}{\textbf{Test Set}} & \multirow{2}{*}{\textbf{Wiki Set}} & \multirow{2}{*}{\textbf{Human Set}} \\ \cmidrule(lr){2-3} \cmidrule(lr){4-5} \cmidrule(lr){6-7}
        & Entity & Query & Entity & Query & Entity & Query &  & \\ \midrule
        \# SEEN entities  & 7,943 & 2,470& 1,604& 199& 7,943&	2,339 & 8,733 & 2,015 \\
        \# SEEN examples & 4,464,748 & 23,514& 51,906& 588& 291,327& 7,460 & 8,733 & 12,057 \\
        \# UNSEEN entities & 0 & 0& 1,588  & 433& 7,944&	3,096 & 1,956,412 & 2,429 \\
        \# UNSEEN examples & 0 & 0& 56,549& 1,406& 316,817& 7,979 & 1,956,412 & 11,100 \\
        \# Total examples  & 4,464,748 & 23,514& 108,455& 1,964& 608,144& 15,439 & 1,965,145 & 23,157\\ \bottomrule
    \end{tabular}
    \caption{Statistics of the Mask\textsc{Oven}-Wiki.}
    \label{tab:overview}
\end{table*}

\paragraph{Entity Distribution.} Figure \ref{fig:distribution_a} shows the distribution of categories in the Mask\textsc{Oven}-Wiki dataset. We have identified 10 primary categories and grouped less prevalent categories under the `others' category. Figure \ref{fig:comparison} shows a more detailed distribution with numbers of each category both in the \textsc{Oven}-Wiki and Mask\textsc{Oven}-Wiki. As shown in \cref{fig:comparison}, we note that the highest proportion of unannotated entities is found in the location, building, and sports categories. Entities in these categories may hit the first and third data filtering rules and be dropped.

\paragraph{Visual Mention Distribution.} Figure \ref{fig:ratios} shows a histogram of the area ratio of visual mentions in images, computed as $a_m / a_i$, where $a_m$ and $a_i$ represent the area of the mention and the image, respectively. The distribution exhibits a generally smooth profile, with an increase in frequency when the area ratio surpasses 95\%, which is primarily caused by the third filtering rule.

\section{Method}
\subsection{Model Architecture}
\begin{figure}
    \centering
    \includegraphics[width=\columnwidth]{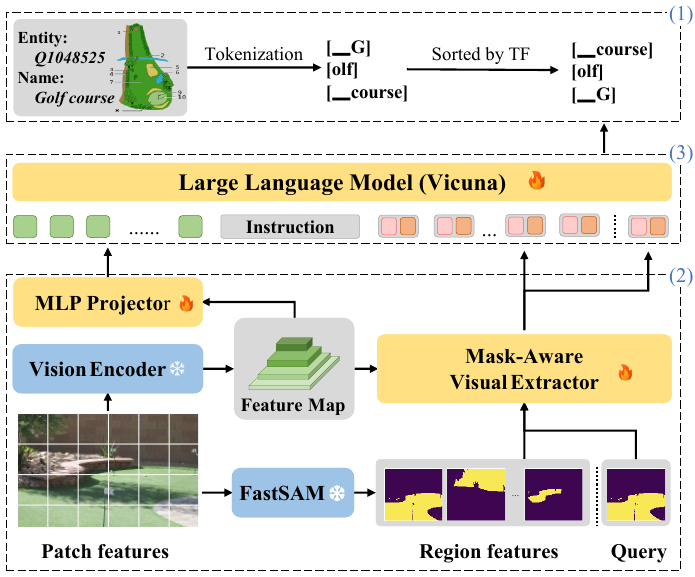}
    \caption{Model overview including 1) pre-built ALD codes for entities, 2) visual semantic tokenization, 3) autoregressive decoding target entity codes. Yellow denotes trainable parameters and blue denotes frozen parameters}
    \label{fig:model}
\end{figure}

Figure \ref{fig:model} illustrates our model overview. We employ visual instruction tuning to train the MLLM in autoregressively decoding the pre-constructed target entity ALD code. Following the generative entity recognition framework of GER-ALD \cite{caron2024generative}, we construct the ALD code for entity $e \in \mathcal{K}$ as
\begin{equation}
\label{eq:ald}
\begin{aligned}
   \text{ALD}_{e} &= \mathcal{S}^{L}\bigl( \mathcal{T}^{T}(e), \bigcup_{e_i \in \mathcal{K}} \mathcal{T}^{T}(e_i)\bigr)
\end{aligned}
\end{equation}
Where $\mathcal{T}^{T}$ is the text tokenizer of LLM, and $\mathcal{S}^{L}$ denotes a function taking the first $L$ tokens in ascending order of term frequency. $L$ denotes the ALD code length. LLM autoregressively generates $\text{ALD}_{e}$ with embedding matrix $\mathbf{Y}$, instruction $\mathbf{X}_{ins}$, image $I$'s features $\mathbf{X}_{I}$ and mask query embedding $\mathbf{X}_{m}$ as follows
\begin{equation}
    \text{ALD}^{\hat e}_{i} = \text{LLM}\bigl(\mathbf{X}_{ins},\mathbf{X}_{I}, \mathbf{X}_{m}, \mathbf{Y}_{\text{ALD}^{\hat e}_{0 \leq j < i}}\bigr)
\end{equation}

Our backbone is based on Osprey\cite{yuanOspreyPixelUnderstanding2023}, a pixel-level MLLM designed for general visual understanding. Following Osprey's settings, we employ the ConvNeXt CLIP \cite{liuConvNet2020s2022} as the vision encoder, Vicuna \cite{vicuna2023} as the foundational LLM, and a vision-language projector using a multilayer perceptron (MLP). Additionally, we reuse its mask-aware visual extractor for constructing regional-level features.

Our method utilizes visual semantic tokenization to extract the fine-grained semantic features from images. It achieves this by reusing feature maps from the vision encoder and parameters from the mask-aware visual extractor, enabling minimal computational and parameter overhead.

\subsection{Visual Semantic Tokenization for Region-Interacted Attention}

Current MLLMs \cite{liuVisualInstructionTuning2023, yuanOspreyPixelUnderstanding2023} use vision encoders like ViT \cite{dosovitskiyImageWorth16x162020} or ResNet \cite{He_2016_CVPR}. These encoders tokenize images based on spatial location rather than semantic content, so that the visual tokens contain incomplete and non-independent semantics, and require additional cross-modal projectors. While Osprey \cite{yuanOspreyPixelUnderstanding2023} and GLaMM \cite{rasheedGLaMMPixelGrounding2023} use region encoders to represent user-specified regions, they do not enhance overall image understanding. PL-VEL focuses on pixel-level visual understanding, motivating us to tokenize images based on semantic content. This approach aligns the semantic granularity of image tokens with the instruction or entity text tokens by controlling each visual token to represent an object, enabling feature interaction within a unified semantic space.

To achieve this, a SAM-like model, FastSAM \cite{zhao2023fast}, executes ``segment-everything" on the image $I$ as a visual semantic tokenizer $\mathcal{T}^{I}$. Subsequently, the mask-aware visual extractor $\mathcal{M}$ takes the binary mask of the region $r$ and the image $I$ as input, encoding these into two embeddings, $\mathbf{x}^{sem}_{r}$ and $\mathbf{x}^{pos}_{r}$, which correspond to semantic and positional feature, respectively. The region feature set $\mathbf{X}_I^{reg}$ is
\begin{equation}
    \mathbf{X}_I^{reg} = \bigl \{\mathbf{x}^{sem}_{r}, \mathbf{x}^{pos}_{r} = \mathcal{M}(I, r) \; | \; r\in\mathcal{T}^{I}(I) \bigr \}
    \label{eq:region_feature}
\end{equation}
Compared to position-based tokenization, semantic tokenization loses the natural token order. Similar to human visual habits, which typically begin with an overview of larger image areas before concentrating on finer details, we arrange the $\mathbf{X}_I^{reg}$ in descending order based on their area $a_r$. This method emulates the human visual attention habit ensuring that larger areas receive broader attention within the autoregressive framework. Then we concatenate region features with the patch features $\mathbf{X}_I^{pat}$ to form the image feature $\mathbf{X}_{I}$. 
\begin{equation}
    \label{eq:img_feature}
    \mathbf{X}_{I} = \bigl [\mathbf{X}_I^{pat}; (\underbrace{\mathbf{x}_{r_1}, \mathbf{x}_{r_2}, \cdots , \mathbf{x}_{r_{|\mathbf{X}_{I}^{reg}|}}}_{\mathbf{x}_{r} \in \mathbf{X}_I^{reg} \; \land \; a_{r_i} > a_{r_{i+1}}}) \bigr ]
\end{equation}

\subsection{Training}
We have implemented a two-stage training strategy for our model. The vision encoder ConvNeXt CLIP \cite{liuConvNet2020s2022} and the semantic tokenizer FastSAM \cite{zhao2023fast} remain frozen, while the mask-aware visual extractor $\mathcal{M}$ and the visual-language projector are fully fine-tuned. The base LLM is fine-tuned with the LoRA \cite{hu2022lora} approach. Both stages employ autoregressive language modeling loss to predict the next token \cite{NEURIPS2023_6dcf277e}. In the first stage, we pre-train on the wiki split to embed entities from knowledge base $\mathcal{K}$ into the model parameters. In the second stage, we fine-tune the model on the entity and query splits to enhance its capability of fine-grained visual entity linking.
\begin{table*}[!htp]
\centering

\begin{tabular}{llcccccccccc}
\toprule
\multirow{2}{*}{Prompt} & \multicolumn{1}{l}{\multirow{2}{*}{Method}} & \multicolumn{3}{c}{Category} & \multicolumn{3}{c}{Validation} & \multicolumn{4}{c}{Test} \\  \cmidrule(r){3-5} \cmidrule(r){6-8} \cmidrule(r){9-12}
 & \multicolumn{1}{c}{} & $\mathcal{R}$ & $\mathcal{G}$ & $\mathcal{Z}$ & \multicolumn{1}{l}{Entity} & \multicolumn{1}{l}{Query} & \multicolumn{1}{l}{Overall} & \multicolumn{1}{l}{Entity} & \multicolumn{1}{l}{Query} & \multicolumn{1}{l}{Human} & \multicolumn{1}{l}{Overall} \\ \midrule
None & \multicolumn{1}{l|}{CLIP \cite{huOpendomainVisualEntity2023}} & \textcolor[RGB]{70,150,90}{\ding{51}} & \textcolor{red}{\ding{55}} & \multicolumn{1}{c|}{\textcolor{red}{\ding{55}}} & 5.4 & 1.2 & \multicolumn{1}{c|}{5.2} & 5.3 & 1.6 & 5.2 & 5.2 \\ \midrule
\multirow{6}{*}{Text} & \multicolumn{1}{l|}{CLIP Fusion \cite{huOpendomainVisualEntity2023}} & \textcolor[RGB]{70,150,90}{\ding{51}} & \textcolor{red}{\ding{55}} & \multicolumn{1}{c|}{\textcolor{red}{\ding{55}}} & 19.0 & 11.9 & \multicolumn{1}{c|}{18.8} & 19.2 & 14.5 & 11.4 & 18.9 \\
 & \multicolumn{1}{l|}{CLIP2CLIP \cite{huOpendomainVisualEntity2023}} & \textcolor[RGB]{70,150,90}{\ding{51}} & \textcolor{red}{\ding{55}} & \multicolumn{1}{c|}{\textcolor{red}{\ding{55}}} & 11.4 & 2.8 & \multicolumn{1}{c|}{11.2} & 11.6 & 3.5 & 12.7 & 11.4 \\
 & \multicolumn{1}{l|}{PaLI-3B \cite{huOpendomainVisualEntity2023}} & \textcolor{red}{\ding{55}} & \textcolor[RGB]{70,150,90}{\ding{51}} & \multicolumn{1}{c|}{\textcolor{red}{\ding{55}}} & 14.3 & 20.5 & \multicolumn{1}{c|}{14.5} & 12.6 & 20.3 & 24.1 & 13.2 \\
 & \multicolumn{1}{l|}{PaLI-17B \cite{huOpendomainVisualEntity2023}} & \textcolor{red}{\ding{55}} & \textcolor[RGB]{70,150,90}{\ding{51}} & \multicolumn{1}{c|}{\textcolor{red}{\ding{55}}} & 21.8 & 29.2 & \multicolumn{1}{c|}{22.0} & 19.8 & 29.5 & 34.1 & 20.5 \\
 & \multicolumn{1}{l|}{BLIP-2 \cite{xiaoGroundingLanguageModels2024a}} & \textcolor{red}{\ding{55}} & \textcolor[RGB]{70,150,90}{\ding{51}} & \multicolumn{1}{c|}{\textcolor[RGB]{70,150,90}{\ding{51}}} & 6.1 & 19.8 &\multicolumn{1}{c|}{6.4} & - & - & - & - \\
 & \multicolumn{1}{l|}{GPT-4V \cite{xiaoGroundingLanguageModels2024a}} & \textcolor{red}{\ding{55}} & \textcolor[RGB]{70,150,90}{\ding{51}} & \multicolumn{1}{c|}{\textcolor[RGB]{70,150,90}{\ding{51}}} & 24.7 & 53.9 & \multicolumn{1}{c|}{25.5} & - & - & - & - \\ \midrule
\multirow{1}{*}{Box} & \multicolumn{1}{l|}{GlaMM} & \textcolor{red}{\ding{55}} & \textcolor[RGB]{70,150,90}{\ding{51}} & \multicolumn{1}{c|}{\textcolor[RGB]{70,150,90}{\ding{51}}} & 1.4 & 8.9 & \multicolumn{1}{c|}{1.6} & 1.5 & 6.1 & 4.3 & 1.7 \\ \midrule
 \multirow{3}{*}{Mask} & \multicolumn{1}{l|}{Osprey-7B} & \textcolor{red}{\ding{55}} & \textcolor[RGB]{70,150,90}{\ding{51}} & \multicolumn{1}{c|}{\textcolor[RGB]{70,150,90}{\ding{51}}} & 0.6 & 9.8 & \multicolumn{1}{c|}{0.8} & 1.0 & 8.2 & 5.6 & 1.3 \\
 & \multicolumn{1}{l|}{Osprey-7B-FT} & \textcolor{red}{\ding{55}} & \textcolor[RGB]{70,150,90}{\ding{51}} & \multicolumn{1}{c|}{\textcolor{red}{\ding{55}}} & 19.4 & 8.3 & \multicolumn{1}{c|}{19.0} & 20.1 & 11.8 & 23.2 & 20.0 \\
 & \multicolumn{1}{l|}{\cellcolor{lightgray}Osprey-Seg-7B} & \cellcolor{lightgray}\textcolor{red}{\ding{55}}  & \cellcolor{lightgray}\textcolor[RGB]{70,150,90}{\ding{51}} & \multicolumn{1}{c|}{\cellcolor{lightgray}\textcolor{red}{\ding{55}}} & \cellcolor{lightgray}24.3 & \cellcolor{lightgray}11.8 & \multicolumn{1}{c|}{\cellcolor{lightgray}24.0} & \cellcolor{lightgray}25.4 & \cellcolor{lightgray}16.1 & \cellcolor{lightgray}25.9 & \cellcolor{lightgray}25.2 \\ \bottomrule
\end{tabular}

\caption{Comparison of VEL models on \textsc{Oven}-Wiki (Text, None) and Mask\textsc{Oven}-Wiki (Mask, Box) validation and test sets. Method categories are denoted as follows: $\mathcal{R}$ for retrieval-based discriminative models, $\mathcal{G}$ for generative models, and $\mathcal{Z}$ for zero-shot models without fine-tuning. The \sethlcolor{lightgray}\hl{gray} line highlights our proposed method.}

\label{tab:main-result}
\end{table*}
\section{Experiments}

\subsection{Experimental Setting}
\paragraph{Metrics.} We evaluate model performance on the validation and test sets of Mask\textsc{Oven}-Wiki using accuracy as the primary metric. Accuracy is computed for the entity and query splits, as well as the human set (test only). To address the challenges zero-shot models face in generating ALD codes and valid entity names, we use BM25 to search the 6 million Wikipedia entity names and take the top-1 result as the prediction.

\paragraph{Data Processing.} 
The pre-train stage used about 2 million wiki split samples. Due to computational resource constraints and the large size of the dataset (approximately 4.5 million samples), we limited the number of annotated samples per entity to fewer than 50 during the fine-tuning stage. As a result, we used about 7\% of the total samples (approximately 0.3 million) in the fine-tuning stage. In addition, all input images were uniformly preprocessed to 512 $\times$ 512. The length of the ALD code is limited to 4 tokens.

\subsection{Main Results}
In \cref{tab:main-result}, we compare the results of VEL models based on different types of prompts in the validation and test sets of \textsc{Oven}-Wiki \cite{huOpendomainVisualEntity2023} (Text) and Mask\textsc{Oven}-Wiki (Mask). Where the ``None" prompt denotes that no prompt was utilized to reference the visual mention. Text-based results are from \citet{huOpendomainVisualEntity2023} and \citet{xiaoGroundingLanguageModels2024a}.

\paragraph{Effectiveness of Mask\textsc{Oven}-Wiki.} In the box and mask prompts, $\mathcal{Z}$ denotes whether the result has been fine-tuned using our dataset. Osprey-7B \cite{yuanOspreyPixelUnderstanding2023} achieves 1.3\% in the zero-shot setting and 20.0\% after fine-tuning, demonstrating the usefulness of our dataset. By introducing visual semantic tokenization, Osprey-7B-Seg improves the overall performance by 3.4\% on the validation set and 5.2\% on the test set.

\paragraph{Advances of Pixel Mask Reference.} Results in \cref{tab:main-result} verify the advantages compared with text and box. Compared with text-based results (6.4\%-25.5\%), our mask representation methods achieve similar performance (0.8\%-25.2\%), despite text prompts offering more detailed descriptions. Compared with box results (around 1.6\%), mask prompts achieve better results. Additionally, we analyzed the limitations of mask methods when dealing with query split, where some questions include additional intents (e.g. ``made of", ``produced by") from original VQA datasets. These situations fall outside the scope of VEL.

\subsection{Analysis and Ablation Study}
\paragraph{Direct versus Reverse Process.} Comparing the experimental results in tables \ref{tab:anno_evaluation} and \ref{tab:main-result}, we observe a performance gap between the direct PL-VEL methods and reverse annotation approaches. GPT-4V achieves an accuracy of 25.5\% in the direct setting. The reverse annotation process, which is an open-vocabulary segmentation task, achieves an accuracy of 94.8\%. These findings show the usefulness of our proposed reverse annotation approach for the PL-VEL task.

\paragraph{Semantic Tokenization and Training.} The ablation experiments evaluate the effectiveness of visual semantic tokenization and training in \cref{tab:pt_ft}. The results indicate that the introduction of region features improves model accuracy in the entity split by 3.7\% to 5.0\% and in the query split by 3.5\% to 5.5\%. In addition, fine-tuning improves the accuracy of the model, whereas the impact of pre-training is relatively limited, with improvements ranging from 0.1\% to 1.6\%. This finding contrasts with those of GER-ALD \cite{caron2024generative}. We attribute the success of GER-ALD's pre-training to its larger pre-training dataset (Entity-WebLI, 55M) and the lighter model (GIT, 0.4B) \cite{wang2022git}.

\begin{table}[ht]
\centering
    \begin{tabular}{llll}
    \toprule
    Method & Entity & Query & Overall \\ \midrule
    Osprey-7B & 0.6 & 7.7 & 0.8 \\
    \quad$\text{}_{\text{+FT}}$ & 19.0 & 6.2 & 18.7 \\
    \quad$\text{}_{\text{+FT +Seg}}$ & \underline{22.7}\small{ +3.7} & \underline{11.7}\small{ +5.5} & \underline{22.4}\small{ +3.7} \\
    \quad$\text{}_{\text{+PT +FT}}$  & 19.3 & 8.3 & 19.0 \\
    \quad$\text{}_{\text{+PT +FT +Seg}}$  & \textbf{24.3}\small{ +5.0} & \textbf{11.8}\small{ +3.5} & \textbf{24.0}\small{ +5.0} \\ \bottomrule
    \end{tabular}
\caption{Ablation study on the validation dataset. PT refers to pre-training, FT refers to fine-tuning, and Seg represents visual semantic tokenization. Bold indicates the best results, and underline denotes the second-best results.}
\label{tab:pt_ft}
\end{table}

\paragraph{Retrieval versus Generation.} Table \ref{tab:ra_result} compares retrieval-based and generation-based methods. PL-VEL is a newly introduced task, so we primarily compare our results with text-based reference models. The absence of handcrafted text queries may create a disadvantage for our approach. \textsc{Auto}VER \cite{xiaoGroundingLanguageModels2024a} is a recently proposed text-based VEL model and demonstrates approximately an 18\% improvement in performance by combining retrieval augmentation ($\mathcal{R}$) and generative prediction ($\mathcal{G}$). Notably, $\text{\textsc{AutoVER}-7B}_\text{p}$ is a peer version of \textsc{Auto}VER model without retrieval augmentation, and its performance closely matches ours (-0.5\%). This finding indicates that retrieval augmentation has the potential to benefit the PL-VEL task.

\begin{table}[ht]
\centering
    \begin{tabular}{lccccc}
    \toprule
    \multirow{2}{*}{Method} & \multicolumn{2}{c}{Type} & \multicolumn{3}{c}{Dev} \\ \cmidrule(r){2-3} \cmidrule(r){4-6}
    \multicolumn{1}{c}{} & $\mathcal{R}$ & $\mathcal{G}$ & Entity & Query & Overall \\ \midrule
    \multicolumn{1}{l|}{CLIP Fusion} & \textcolor[RGB]{70,150,90}{\ding{51}} & \multicolumn{1}{c|}{\textcolor{red}{\ding{55}}} & 19.0 & 11.9 & 18.8  \\
    \multicolumn{1}{l|}{PaLI-17B} & \textcolor{red}{\ding{55}} & \multicolumn{1}{c|}{\textcolor[RGB]{70,150,90}{\ding{51}}} & 21.8 & 29.2 & 22.0 \\
    \multicolumn{1}{l|}{\textsc{AutoVER}-7B} & \textcolor[RGB]{70,150,90}{\ding{51}} & \multicolumn{1}{c|}{\textcolor[RGB]{70,150,90}{\ding{51}}} & 42.2 & 43.1 & 42.3 \\
    \multicolumn{1}{l|}{\textsc{AutoVER}-13B} & \textcolor[RGB]{70,150,90}{\ding{51}} & \multicolumn{1}{c|}{\textcolor[RGB]{70,150,90}{\ding{51}}} & 44.7 & 43.6 & 44.6 \\
    \multicolumn{1}{l|}{\cellcolor{lightgray}$\textsc{AutoVER-7B}_{\text{p}}$} & \cellcolor{lightgray}\textcolor{red}{\ding{55}} & \multicolumn{1}{c|}{\cellcolor{lightgray}\textcolor[RGB]{70,150,90}{\ding{51}}} & \cellcolor{lightgray}23.8 & \cellcolor{lightgray}- & \cellcolor{lightgray}- \\ \midrule
    \multicolumn{1}{l|}{Osprey-Seg-7B} & \textcolor{red}{\ding{55}} & \multicolumn{1}{c|}{\textcolor[RGB]{70,150,90}{\ding{51}}} & 24.3 & 11.8 & 24.0 \\ \bottomrule
    \end{tabular}
\caption{Comparing the retrieval-based and generation-based methods on the validation dataset. \sethlcolor{lightgray}\hl{Gray} line represents peer results of \textsc{AutoVER}.}
\label{tab:ra_result}
\end{table}

\section{Conclusion}
In this paper, we introduce the Pixel-Level Visual Entity Linking (PL-VEL) task, which links visual mentions indicated by pixel masks to entities in a knowledge base. This task is a supplement to the text-based VEL, enhancing VEL's practicality for tasks like VQA, visual reasoning, and detailed image captioning. We developed the Mask\textsc{Oven}-Wiki dataset, a multimodal dataset aligning pixel-level regions with entity-level labels, achieving 94.8\% annotation accuracy. Models trained on this dataset achieved over an 18-point improvement in accuracy compared to zero-shot models, with our visual semantic tokenization method contributing an additional 5-point increase. Despite these gains, the final model's linking accuracy was about 25\%, indicating both the effectiveness of reverse annotation and the potential of the Mask\textsc{Oven}-Wiki dataset for enabling fine-grained visual understanding in MLLMs.

\section*{Acknowledgments}
This work is supported by the National Natural Science Foundation of China (No. 62172044). We thank the anonymous reviewers for their kind comments.
\bibliography{aaai25}

\begin{thebibliography}{54}
\providecommand{\natexlab}[1]{#1}

\bibitem[{Bossard, Guillaumin, and Van~Gool(2014)}]{food101}
Bossard, L.; Guillaumin, M.; and Van~Gool, L. 2014.
\newblock Food-101 -- Mining Discriminative Components with Random Forests.
\newblock In Fleet, D.; Pajdla, T.; Schiele, B.; and Tuytelaars, T., eds., \emph{Computer Vision -- ECCV 2014}, 446--461. Cham: Springer International Publishing.
\newblock ISBN 978-3-319-10599-4.

\bibitem[{Caron et~al.(2024{\natexlab{a}})Caron, Iscen, Fathi, and Schmid}]{caronGenerativeApproachWikipediaScale2024}
Caron, M.; Iscen, A.; Fathi, A.; and Schmid, C. 2024{\natexlab{a}}.
\newblock A {{Generative Approach}} for {{Wikipedia-Scale Visual Entity Recognition}}.
\newblock arxiv:2403.02041.

\bibitem[{Caron et~al.(2024{\natexlab{b}})Caron, Iscen, Fathi, and Schmid}]{caron2024generative}
Caron, M.; Iscen, A.; Fathi, A.; and Schmid, C. 2024{\natexlab{b}}.
\newblock A Generative Approach for Wikipedia-Scale Visual Entity Recognition.
\newblock In \emph{Proceedings of the IEEE/CVF Conference on Computer Vision and Pattern Recognition}, 17313--17322.

\bibitem[{Chang et~al.(2024)Chang, Bao, Hou, Jiang, Zheng, Zhong, Zhang, Song, Yao, Jiang, Lin, Jin, and Liu}]{chang2024fluxfastsoftwarebasedcommunication}
Chang, L.-W.; Bao, W.; Hou, Q.; Jiang, C.; Zheng, N.; Zhong, Y.; Zhang, X.; Song, Z.; Yao, C.; Jiang, Z.; Lin, H.; Jin, X.; and Liu, X. 2024.
\newblock FLUX: Fast Software-based Communication Overlap On GPUs Through Kernel Fusion.
\newblock arXiv:2406.06858.

\bibitem[{Chen and Wu(2024)}]{Chen_2024_CVPR}
Chen, K.; and Wu, X. 2024.
\newblock VTQA: Visual Text Question Answering via Entity Alignment and Cross-Media Reasoning.
\newblock In \emph{Proceedings of the IEEE/CVF Conference on Computer Vision and Pattern Recognition (CVPR)}, 27218--27227.

\bibitem[{Chen et~al.(2023)Chen, Zhang, Zeng, Zhang, Zhu, and Zhao}]{chenShikraUnleashingMultimodal2023}
Chen, K.; Zhang, Z.; Zeng, W.; Zhang, R.; Zhu, F.; and Zhao, R. 2023.
\newblock Shikra: {{Unleashing Multimodal LLM}}'s {{Referential Dialogue Magic}}.
\newblock arxiv:2306.15195.

\bibitem[{Chiang et~al.(2023)Chiang, Li, Lin, Sheng, Wu, Zhang, Zheng, Zhuang, Zhuang, Gonzalez, Stoica, and Xing}]{vicuna2023}
Chiang, W.-L.; Li, Z.; Lin, Z.; Sheng, Y.; Wu, Z.; Zhang, H.; Zheng, L.; Zhuang, S.; Zhuang, Y.; Gonzalez, J.~E.; Stoica, I.; and Xing, E.~P. 2023.
\newblock Vicuna: An Open-Source Chatbot Impressing GPT-4 with 90\%* ChatGPT Quality.

\bibitem[{Deng et~al.(2009)Deng, Dong, Socher, Li, Li, and Fei-Fei}]{deng2009imagenet}
Deng, J.; Dong, W.; Socher, R.; Li, L.-J.; Li, K.; and Fei-Fei, L. 2009.
\newblock Imagenet: A large-scale hierarchical image database.
\newblock In \emph{2009 IEEE conference on computer vision and pattern recognition}, 248--255. Ieee.

\bibitem[{Dosovitskiy et~al.(2020)Dosovitskiy, Beyer, Kolesnikov, Weissenborn, Zhai, Unterthiner, Dehghani, Minderer, Heigold, Gelly, Uszkoreit, and Houlsby}]{dosovitskiyImageWorth16x162020}
Dosovitskiy, A.; Beyer, L.; Kolesnikov, A.; Weissenborn, D.; Zhai, X.; Unterthiner, T.; Dehghani, M.; Minderer, M.; Heigold, G.; Gelly, S.; Uszkoreit, J.; and Houlsby, N. 2020.
\newblock An {{Image}} Is {{Worth}} 16x16 {{Words}}: {{Transformers}} for {{Image Recognition}} at {{Scale}}.
\newblock In \emph{International {{Conference}} on {{Learning Representations}}}.

\bibitem[{Goyal et~al.(2017)Goyal, Khot, Summers-Stay, Batra, and Parikh}]{8100153}
Goyal, Y.; Khot, T.; Summers-Stay, D.; Batra, D.; and Parikh, D. 2017.
\newblock Making the V in VQA Matter: Elevating the Role of Image Understanding in Visual Question Answering.
\newblock In \emph{2017 IEEE Conference on Computer Vision and Pattern Recognition (CVPR)}, 6325--6334.

\bibitem[{Guo et~al.(2024)Guo, De~Mello, Yin, Byeon, Cheung, Yu, Luo, and Liu}]{guo2024regiongpt}
Guo, Q.; De~Mello, S.; Yin, H.; Byeon, W.; Cheung, K.~C.; Yu, Y.; Luo, P.; and Liu, S. 2024.
\newblock Regiongpt: Towards region understanding vision language model.
\newblock In \emph{Proceedings of the IEEE/CVF Conference on Computer Vision and Pattern Recognition}, 13796--13806.

\bibitem[{He et~al.(2016)He, Zhang, Ren, and Sun}]{He_2016_CVPR}
He, K.; Zhang, X.; Ren, S.; and Sun, J. 2016.
\newblock Deep Residual Learning for Image Recognition.
\newblock In \emph{Proceedings of the IEEE Conference on Computer Vision and Pattern Recognition (CVPR)}.

\bibitem[{Hu et~al.(2022)Hu, yelong shen, Wallis, Allen-Zhu, Li, Wang, Wang, and Chen}]{hu2022lora}
Hu, E.~J.; yelong shen; Wallis, P.; Allen-Zhu, Z.; Li, Y.; Wang, S.; Wang, L.; and Chen, W. 2022.
\newblock Lo{RA}: Low-Rank Adaptation of Large Language Models.
\newblock In \emph{International Conference on Learning Representations}.

\bibitem[{Hu et~al.(2023)Hu, Luan, Chen, Khandelwal, Joshi, Lee, Toutanova, and Chang}]{huOpendomainVisualEntity2023}
Hu, H.; Luan, Y.; Chen, Y.; Khandelwal, U.; Joshi, M.; Lee, K.; Toutanova, K.; and Chang, M.-W. 2023.
\newblock Open-Domain {{Visual Entity Recognition}}: {{Towards Recognizing Millions}} of {{Wikipedia Entities}}.
\newblock In \emph{Proceedings of the {{IEEE}}/{{CVF International Conference}} on {{Computer Vision}}}, 12065--12075.

\bibitem[{Huang et~al.(2024)Huang, Zhang, Ma, Tian, Feng, Zhang, Li, Guo, and Zhang}]{huang2024tagtext}
Huang, X.; Zhang, Y.; Ma, J.; Tian, W.; Feng, R.; Zhang, Y.; Li, Y.; Guo, Y.; and Zhang, L. 2024.
\newblock Tag2Text: Guiding Vision-Language Model via Image Tagging.
\newblock In \emph{The Twelfth International Conference on Learning Representations}.

\bibitem[{Kirillov et~al.(2023)Kirillov, Mintun, Ravi, Mao, Rolland, Gustafson, Xiao, Whitehead, Berg, Lo, Doll{\'a}r, and Girshick}]{kirillovSegmentAnything2023}
Kirillov, A.; Mintun, E.; Ravi, N.; Mao, H.; Rolland, C.; Gustafson, L.; Xiao, T.; Whitehead, S.; Berg, A.~C.; Lo, W.-Y.; Doll{\'a}r, P.; and Girshick, R. 2023.
\newblock Segment {{Anything}}.
\newblock arxiv:2304.02643.

\bibitem[{Krause et~al.(2013)Krause, Stark, Deng, and Fei-Fei}]{6755945}
Krause, J.; Stark, M.; Deng, J.; and Fei-Fei, L. 2013.
\newblock 3D Object Representations for Fine-Grained Categorization.
\newblock In \emph{2013 IEEE International Conference on Computer Vision Workshops}, 554--561.

\bibitem[{Krishna et~al.(2017)Krishna, Zhu, Groth, Johnson, Hata, Kravitz, Chen, Kalantidis, Li, Shamma, Bernstein, and Fei-Fei}]{vg}
Krishna, R.; Zhu, Y.; Groth, O.; Johnson, J.; Hata, K.; Kravitz, J.; Chen, S.; Kalantidis, Y.; Li, L.-J.; Shamma, D.~A.; Bernstein, M.~S.; and Fei-Fei, L. 2017.
\newblock Visual Genome: Connecting Language and Vision Using Crowdsourced Dense Image Annotations.
\newblock \emph{Int. J. Comput. Vision}, 123(1): 32–73.

\bibitem[{Li et~al.(2022)Li, Li, Xiong, and Hoi}]{liBLIPBootstrappingLanguageImage2022}
Li, J.; Li, D.; Xiong, C.; and Hoi, S. 2022.
\newblock {{BLIP}}: {{Bootstrapping Language-Image Pre-training}} for {{Unified Vision-Language Understanding}} and {{Generation}}.
\newblock In \emph{Proceedings of the 39th {{International Conference}} on {{Machine Learning}}}, 12888--12900. PMLR.

\bibitem[{Lin et~al.(2014)Lin, Maire, Belongie, Bourdev, Girshick, Hays, Perona, Ramanan, Doll{'{a} }r, and Zitnick}]{cocodataset}
Lin, T.; Maire, M.; Belongie, S.~J.; Bourdev, L.~D.; Girshick, R.~B.; Hays, J.; Perona, P.; Ramanan, D.; Doll{'{a} }r, P.; and Zitnick, C.~L. 2014.
\newblock Microsoft {COCO:} Common Objects in Context.
\newblock \emph{CoRR}, abs/1405.0312.

\bibitem[{Liu et~al.(2023{\natexlab{a}})Liu, Li, Wu, and Lee}]{NEURIPS2023_6dcf277e}
Liu, H.; Li, C.; Wu, Q.; and Lee, Y.~J. 2023{\natexlab{a}}.
\newblock Visual {{Instruction Tuning}}.
\newblock In Oh, A.; Naumann, T.; Globerson, A.; Saenko, K.; Hardt, M.; and Levine, S., eds., \emph{Advances in Neural Information Processing Systems}, volume~36, 34892--34916. Curran Associates, Inc.

\bibitem[{Liu et~al.(2023{\natexlab{b}})Liu, Li, Wu, and Lee}]{liuVisualInstructionTuning2023}
Liu, H.; Li, C.; Wu, Q.; and Lee, Y.~J. 2023{\natexlab{b}}.
\newblock Visual {{Instruction Tuning}}.
\newblock arxiv:2304.08485.

\bibitem[{Liu et~al.(2023{\natexlab{c}})Liu, Zeng, Ren, Li, Zhang, Yang, Li, Yang, Su, Zhu, and Zhang}]{liuGroundingDINOMarrying2023}
Liu, S.; Zeng, Z.; Ren, T.; Li, F.; Zhang, H.; Yang, J.; Li, C.; Yang, J.; Su, H.; Zhu, J.; and Zhang, L. 2023{\natexlab{c}}.
\newblock Grounding {{DINO}}: {{Marrying DINO}} with {{Grounded Pre-Training}} for {{Open-Set Object Detection}}.
\newblock arxiv:2303.05499.

\bibitem[{Liu et~al.(2023{\natexlab{d}})Liu, Zeng, Ren, Li, Zhang, Yang, Li, Yang, Su, Zhu et~al.}]{liu2023grounding}
Liu, S.; Zeng, Z.; Ren, T.; Li, F.; Zhang, H.; Yang, J.; Li, C.; Yang, J.; Su, H.; Zhu, J.; et~al. 2023{\natexlab{d}}.
\newblock Grounding dino: Marrying dino with grounded pre-training for open-set object detection.
\newblock \emph{arXiv preprint arXiv:2303.05499}.

\bibitem[{Liu et~al.(2022)Liu, Mao, Wu, Feichtenhofer, Darrell, and Xie}]{liuConvNet2020s2022}
Liu, Z.; Mao, H.; Wu, C.-Y.; Feichtenhofer, C.; Darrell, T.; and Xie, S. 2022.
\newblock A ConvNet for the 2020s.
\newblock In \emph{Proceedings of the IEEE/CVF Conference on Computer Vision and Pattern Recognition (CVPR)}, 11976--11986.

\bibitem[{Loshchilov and Hutter(2019)}]{adamw}
Loshchilov, I.; and Hutter, F. 2019.
\newblock Decoupled Weight Decay Regularization.
\newblock In \emph{International Conference on Learning Representations}.

\bibitem[{Maji et~al.(2013)Maji, Rahtu, Kannala, Blaschko, and Vedaldi}]{maji2013finegrainedvisualclassificationaircraft}
Maji, S.; Rahtu, E.; Kannala, J.; Blaschko, M.; and Vedaldi, A. 2013.
\newblock Fine-Grained Visual Classification of Aircraft.
\newblock arXiv:1306.5151.

\bibitem[{Marino et~al.(2019)Marino, Rastegari, Farhadi, and Mottaghi}]{8953725}
Marino, K.; Rastegari, M.; Farhadi, A.; and Mottaghi, R. 2019.
\newblock OK-VQA: A Visual Question Answering Benchmark Requiring External Knowledge.
\newblock In \emph{2019 IEEE/CVF Conference on Computer Vision and Pattern Recognition (CVPR)}, 3190--3199.

\bibitem[{Nilsback and Zisserman(2008)}]{4756141}
Nilsback, M.-E.; and Zisserman, A. 2008.
\newblock Automated Flower Classification over a Large Number of Classes.
\newblock In \emph{2008 Sixth Indian Conference on Computer Vision, Graphics \& Image Processing}, 722--729.

\bibitem[{Peng et~al.(2023)Peng, Wang, Dong, Hao, Huang, Ma, and Wei}]{pengKosmos2GroundingMultimodal2023a}
Peng, Z.; Wang, W.; Dong, L.; Hao, Y.; Huang, S.; Ma, S.; and Wei, F. 2023.
\newblock Kosmos-2: {{Grounding Multimodal Large Language Models}} to the {{World}}.
\newblock arxiv:2306.14824.

\bibitem[{Piosenka(2021)}]{sport100}
Piosenka, G. 2021.
\newblock Sports100: 100 sports image classification.
\newblock \url{https://www.kaggle.com/datasets/gpiosenka/sports-classification}.
\newblock Accessed: 2022-09-26.

\bibitem[{Qiu et~al.(2024)Qiu, Madotto, Lin, Crook, Xu, Dong, Faloutsos, Li, Damavandi, and Moon}]{qiuSnapNTellEnhancingEntityCentric2024a}
Qiu, J.; Madotto, A.; Lin, Z.; Crook, P.~A.; Xu, Y.~E.; Dong, X.~L.; Faloutsos, C.; Li, L.; Damavandi, B.; and Moon, S. 2024.
\newblock {{SnapNTell}}: {{Enhancing Entity-Centric Visual Question Answering}} with {{Retrieval Augmented Multimodal LLM}}.
\newblock arxiv:2403.04735.

\bibitem[{Rasheed et~al.(2023)Rasheed, Maaz, Mullappilly, Shaker, Khan, Cholakkal, Anwer, Xing, Yang, and Khan}]{rasheedGLaMMPixelGrounding2023}
Rasheed, H.; Maaz, M.; Mullappilly, S.~S.; Shaker, A.; Khan, S.; Cholakkal, H.; Anwer, R.~M.; Xing, E.; Yang, M.-H.; and Khan, F.~S. 2023.
\newblock {{GLaMM}}: {{Pixel Grounding Large Multimodal Model}}.
\newblock arxiv:2311.03356.

\bibitem[{Rasheed et~al.(2024)Rasheed, Maaz, Shaji, Shaker, Khan, Cholakkal, Anwer, Xing, Yang, and Khan}]{Rasheed_2024_CVPR}
Rasheed, H.; Maaz, M.; Shaji, S.; Shaker, A.; Khan, S.; Cholakkal, H.; Anwer, R.~M.; Xing, E.; Yang, M.-H.; and Khan, F.~S. 2024.
\newblock GLaMM: Pixel Grounding Large Multimodal Model.
\newblock In \emph{Proceedings of the IEEE/CVF Conference on Computer Vision and Pattern Recognition (CVPR)}, 13009--13018.

\bibitem[{Rasley et~al.(2020)Rasley, Rajbhandari, Ruwase, and He}]{deepspeed}
Rasley, J.; Rajbhandari, S.; Ruwase, O.; and He, Y. 2020.
\newblock DeepSpeed: System Optimizations Enable Training Deep Learning Models with Over 100 Billion Parameters.
\newblock In \emph{Proceedings of the 26th ACM SIGKDD International Conference on Knowledge Discovery \& Data Mining}, KDD '20, 3505–3506. New York, NY, USA: Association for Computing Machinery.
\newblock ISBN 9781450379984.

\bibitem[{Ren et~al.(2024)Ren, Liu, Zeng, Lin, Li, Cao, Chen, Huang, Chen, Yan, Zeng, Zhang, Li, Yang, Li, Jiang, and Zhang}]{renGroundedSAMAssembling2024a}
Ren, T.; Liu, S.; Zeng, A.; Lin, J.; Li, K.; Cao, H.; Chen, J.; Huang, X.; Chen, Y.; Yan, F.; Zeng, Z.; Zhang, H.; Li, F.; Yang, J.; Li, H.; Jiang, Q.; and Zhang, L. 2024.
\newblock Grounded {SAM}: {Assembling} {Open}-{World} {Models} for {Diverse} {Visual} {Tasks}.
\newblock ArXiv:2401.14159 [cs].

\bibitem[{Sain et~al.(2023)Sain, Bhunia, Chowdhury, Koley, Xiang, and Song}]{sainCLIPAllThings2023}
Sain, A.; Bhunia, A.~K.; Chowdhury, P.~N.; Koley, S.; Xiang, T.; and Song, Y.-Z. 2023.
\newblock {{CLIP}} for {{All Things Zero-Shot Sketch-Based Image Retrieval}}, {{Fine-Grained}} or {{Not}}.
\newblock In \emph{Proceedings of the {{IEEE}}/{{CVF Conference}} on {{Computer Vision}} and {{Pattern Recognition}}}, 2765--2775.

\bibitem[{Saito et~al.(2023)Saito, Sohn, Zhang, Li, Lee, Saenko, and Pfister}]{saitoPic2WordMappingPictures2023}
Saito, K.; Sohn, K.; Zhang, X.; Li, C.-L.; Lee, C.-Y.; Saenko, K.; and Pfister, T. 2023.
\newblock {{Pic2Word}}: {{Mapping Pictures}} to {{Words}} for {{Zero-Shot Composed Image Retrieval}}.
\newblock In \emph{Proceedings of the {{IEEE}}/{{CVF Conference}} on {{Computer Vision}} and {{Pattern Recognition}}}, 19305--19314.

\bibitem[{Singh et~al.(2019)Singh, Natarajan, Shah, Jiang, Chen, Batra, Parikh, and Rohrbach}]{8953586}
Singh, A.; Natarajan, V.; Shah, M.; Jiang, Y.; Chen, X.; Batra, D.; Parikh, D.; and Rohrbach, M. 2019.
\newblock Towards VQA Models That Can Read.
\newblock In \emph{2019 IEEE/CVF Conference on Computer Vision and Pattern Recognition (CVPR)}, 8309--8318.

\bibitem[{Sun et~al.(2022)Sun, Fan, Guo, Zhang, and Cheng}]{sunVisualNamedEntity2022}
Sun, W.; Fan, Y.; Guo, J.; Zhang, R.; and Cheng, X. 2022.
\newblock Visual {{Named Entity Linking}}: {{A New Dataset}} and {{A Baseline}}.
\newblock In Goldberg, Y.; Kozareva, Z.; and Zhang, Y., eds., \emph{Findings of the {{Association}} for {{Computational Linguistics}}: {{EMNLP}} 2022}, 2403--2415. Abu Dhabi, United Arab Emirates: Association for Computational Linguistics.

\bibitem[{Van~Horn et~al.(2018)Van~Horn, Mac~Aodha, Song, Cui, Sun, Shepard, Adam, Perona, and Belongie}]{Horn_2018_CVPR}
Van~Horn, G.; Mac~Aodha, O.; Song, Y.; Cui, Y.; Sun, C.; Shepard, A.; Adam, H.; Perona, P.; and Belongie, S. 2018.
\newblock The INaturalist Species Classification and Detection Dataset.
\newblock In \emph{Proceedings of the IEEE Conference on Computer Vision and Pattern Recognition (CVPR)}.

\bibitem[{Wang et~al.(2022)Wang, Yang, Hu, Li, Lin, Gan, Liu, Liu, and Wang}]{wang2022git}
Wang, J.; Yang, Z.; Hu, X.; Li, L.; Lin, K.; Gan, Z.; Liu, Z.; Liu, C.; and Wang, L. 2022.
\newblock {GIT}: A Generative Image-to-text Transformer for Vision and Language.
\newblock \emph{Transactions on Machine Learning Research}.

\bibitem[{Weyand et~al.(2020)Weyand, Araujo, Cao, and Sim}]{weyand2020GLDv2}
Weyand, T.; Araujo, A.; Cao, B.; and Sim, J. 2020.
\newblock {Google Landmarks Dataset v2 - A Large-Scale Benchmark for Instance-Level Recognition and Retrieval}.
\newblock In \emph{Proc. CVPR}.

\bibitem[{Wu et~al.(2023)Wu, Li, Zhao, Wang, Liu, Huai, Yuan, and Chen}]{wuRecognizingUnseenObjects2023}
Wu, L.; Li, Z.; Zhao, H.; Wang, Z.; Liu, Q.; Huai, B.; Yuan, N.~J.; and Chen, E. 2023.
\newblock Recognizing {Unseen} {Objects} via {Multimodal} {Intensive} {Knowledge} {Graph} {Propagation}.
\newblock In \emph{Proceedings of the 29th {ACM} {SIGKDD} {Conference} on {Knowledge} {Discovery} and {Data} {Mining}}, {KDD} '23, 2618--2628. New York, NY, USA: Association for Computing Machinery.
\newblock ISBN 9798400701030.

\bibitem[{Xiao et~al.(2010)Xiao, Hays, Ehinger, Oliva, and Torralba}]{5539970}
Xiao, J.; Hays, J.; Ehinger, K.~A.; Oliva, A.; and Torralba, A. 2010.
\newblock SUN database: Large-scale scene recognition from abbey to zoo.
\newblock In \emph{2010 IEEE Computer Society Conference on Computer Vision and Pattern Recognition}, 3485--3492.

\bibitem[{Xiao et~al.(2024)Xiao, Gong, {Cascante-Bonilla}, Zhang, Wu, and Ordonez}]{xiaoGroundingLanguageModels2024a}
Xiao, Z.; Gong, M.; {Cascante-Bonilla}, P.; Zhang, X.; Wu, J.; and Ordonez, V. 2024.
\newblock Grounding {{Language Models}} for {{Visual Entity Recognition}}.
\newblock arxiv:2402.18695.

\bibitem[{Yuan et~al.(2023)Yuan, Li, Liu, Tang, Luo, Qin, Zhang, and Zhu}]{yuanOspreyPixelUnderstanding2023}
Yuan, Y.; Li, W.; Liu, J.; Tang, D.; Luo, X.; Qin, C.; Zhang, L.; and Zhu, J. 2023.
\newblock Osprey: {{Pixel Understanding}} with {{Visual Instruction Tuning}}.
\newblock https://arxiv.org/abs/2312.10032v2.

\bibitem[{Zhang et~al.(2024{\natexlab{a}})Zhang, Sun, Chen, Xiao, Shao, Zhang, Liu, Chen, and Luo}]{zhangGPT4RoIInstructionTuning2024}
Zhang, S.; Sun, P.; Chen, S.; Xiao, M.; Shao, W.; Zhang, W.; Liu, Y.; Chen, K.; and Luo, P. 2024{\natexlab{a}}.
\newblock {{GPT4RoI}}: {{Instruction Tuning Large Language Model}} on {{Region-of-Interest}}.
\newblock arxiv:2307.03601.

\bibitem[{Zhang et~al.(2024{\natexlab{b}})Zhang, Huang, Ma, Li, Luo, Xie, Qin, Luo, Li, Liu et~al.}]{zhang2024recognize}
Zhang, Y.; Huang, X.; Ma, J.; Li, Z.; Luo, Z.; Xie, Y.; Qin, Y.; Luo, T.; Li, Y.; Liu, S.; et~al. 2024{\natexlab{b}}.
\newblock Recognize anything: A strong image tagging model.
\newblock In \emph{Proceedings of the IEEE/CVF Conference on Computer Vision and Pattern Recognition}, 1724--1732.

\bibitem[{Zhang et~al.(2024{\natexlab{c}})Zhang, Lin, Cao, and Lin}]{zhangEndToEndSpatiallyConstrainedMultiPerspective2024}
Zhang, Y.; Lin, C.; Cao, D.; and Lin, D. 2024{\natexlab{c}}.
\newblock End-{{To-End Spatially-Constrained Multi-Perspective Fine-Grained Image Captioning}}.
\newblock In \emph{{{ICASSP}} 2024 - 2024 {{IEEE International Conference}} on {{Acoustics}}, {{Speech}} and {{Signal Processing}} ({{ICASSP}})}, 3360--3364.

\bibitem[{Zhao et~al.(2023)Zhao, Ding, An, Du, Yu, Li, Tang, and Wang}]{zhao2023fast}
Zhao, X.; Ding, W.; An, Y.; Du, Y.; Yu, T.; Li, M.; Tang, M.; and Wang, J. 2023.
\newblock Fast Segment Anything.
\newblock arXiv:2306.12156.

\bibitem[{Zhu et~al.(2023)Zhu, Chen, Shen, Li, and Elhoseiny}]{zhuMiniGPT4EnhancingVisionLanguage2023a}
Zhu, D.; Chen, J.; Shen, X.; Li, X.; and Elhoseiny, M. 2023.
\newblock {{MiniGPT-4}}: {{Enhancing Vision-Language Understanding}} with {{Advanced Large Language Models}}.
\newblock https://arxiv.org/abs/2304.10592v2.

\bibitem[{Zhu et~al.(2016)Zhu, Groth, Bernstein, and Fei-Fei}]{7780909}
Zhu, Y.; Groth, O.; Bernstein, M.; and Fei-Fei, L. 2016.
\newblock Visual7W: Grounded Question Answering in Images.
\newblock In \emph{2016 IEEE Conference on Computer Vision and Pattern Recognition (CVPR)}, 4995--5004.

\bibitem[{Zou et~al.(2023)Zou, Yang, Zhang, Li, Li, Wang, Wang, Gao, and Lee}]{zouSegmentEverythingEverywhere2023}
Zou, X.; Yang, J.; Zhang, H.; Li, F.; Li, L.; Wang, J.; Wang, L.; Gao, J.; and Lee, Y.~J. 2023.
\newblock Segment {Everything} {Everywhere} {All} at {Once}.
\newblock \emph{Advances in Neural Information Processing Systems}, 36: 19769--19782.

\end{thebibliography}

\newpage
\clearpage
\appendix

\section{More Examples from Mask\textsc{Oven}-Wiki}
\label{append:data_examples}
The Mask\textsc{Oven}-Wiki dataset stores annotations in the COCO \cite{cocodataset} format, which includes details such as image metadata, object categories (entities), and segmentation masks (visual mentions). The masks are encoded using the Run-Length Encoding (RLE) format \cite{cocodataset}, which efficiently represents binary masks by recording the lengths of consecutive runs of pixels.

\begin{figure}[ht]
    \centering
    \includegraphics[width=\columnwidth]{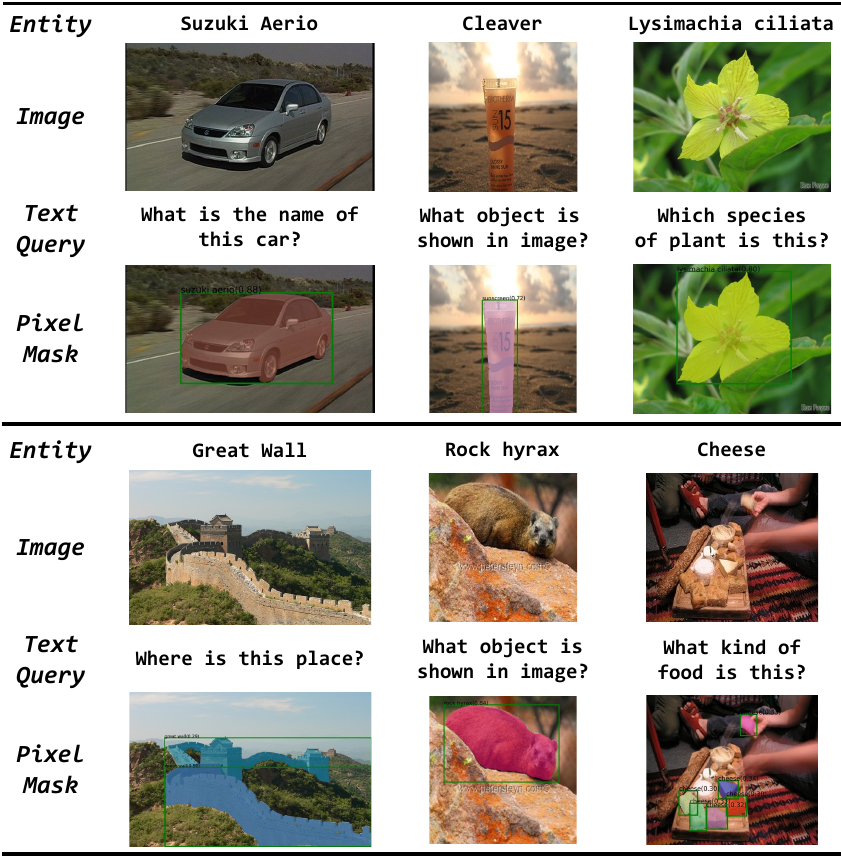}
    \caption{Examples from Mask\textsc{Oven}-Wiki entity split.}
    \label{fig:es-examples}
\end{figure}

\begin{figure}[ht]
    \centering
    \includegraphics[width=\columnwidth]{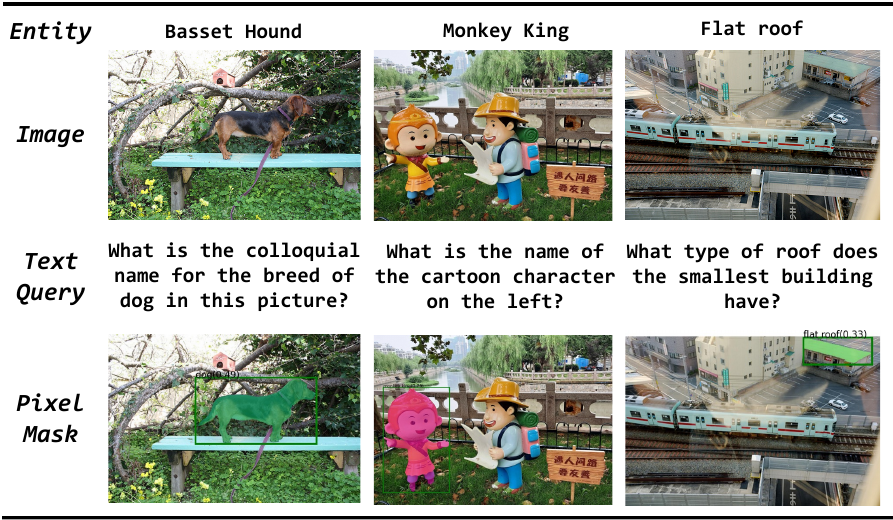}
    \caption{Examples from Mask\textsc{Oven}-Wiki query split.}
    \label{fig:qs-examples}
\end{figure}

To provide more details about the Mask\textsc{Oven}-Wiki dataset, we selected 6 examples from the entity split and 3 examples from the query split. These examples are shown in fig. \ref{fig:es-examples} and \ref{fig:qs-examples}, respectively. The examples try to cover as many different entity types as possible. 

\section{Experiment Details}
\label{append:setup}
\subsection{Annotation Setup}
We utilized a cluster of 30 nodes for the annotation of large-scale data. Each node was configured with 7 CPU cores, 30 GB of memory, and an NVIDIA Tesla P40-24G GPU. For the Mask\textsc{Oven}-Wiki dataset, annotating the Entity split and Query split took approximately 120 hours, while annotating the Wiki split took about 35 hours. The specifications for the annotation models are as follows. SAM \cite{kirillovSegmentAnything2023} used the ViT Huge (ViT-H) version, GroundDINO \cite{liuGroundingDINOMarrying2023} used the Swin-T version, and SEEM \cite{zouSegmentEverythingEverywhere2023} used the Focal-L version. The bounding box threshold was set to 0.3, the text query threshold to 0.25, and the annotation batch size was 3.

\subsection{Experimental Setup}
We conducted the PL-VEL experiments on a machine with 2 NVIDIA A100-40G GPUs. The pre-training parameters were as follows: batch size of 8, gradient accumulation over 2 steps, and 30,000 training steps, which took approximately 157 hours. The learning rate was initially tested with different settings [1e-7, 1e-5, 1e-4, 1e-3] during the first 2,000 steps and was ultimately set at 1e-4. 

The fine-tuning parameters were as follows: batch size of 8, gradient accumulation over 4 steps, and 10,000 training steps that took approximately 48 hours. The learning rate was tested with settings [1e-7, 1e-5, 1e-4] over the first 2,000 steps and was finalized at 1e-4. Due to the large dataset and limited time, we limited the maximum number of samples per entity to 50 during fine-tuning. 

The entire experiment was implemented using PyTorch, with model parameters optimized by the AdamW \cite{adamw} algorithm and data parallel training facilitated by DeepSpeed ZeRO-0 \cite{deepspeed}. The maximum sequence length for the LLM was set to 2048, and the image resolution was scaled to 512 $\times$ 512.
\begin{figure*}
\includegraphics[width=\textwidth]{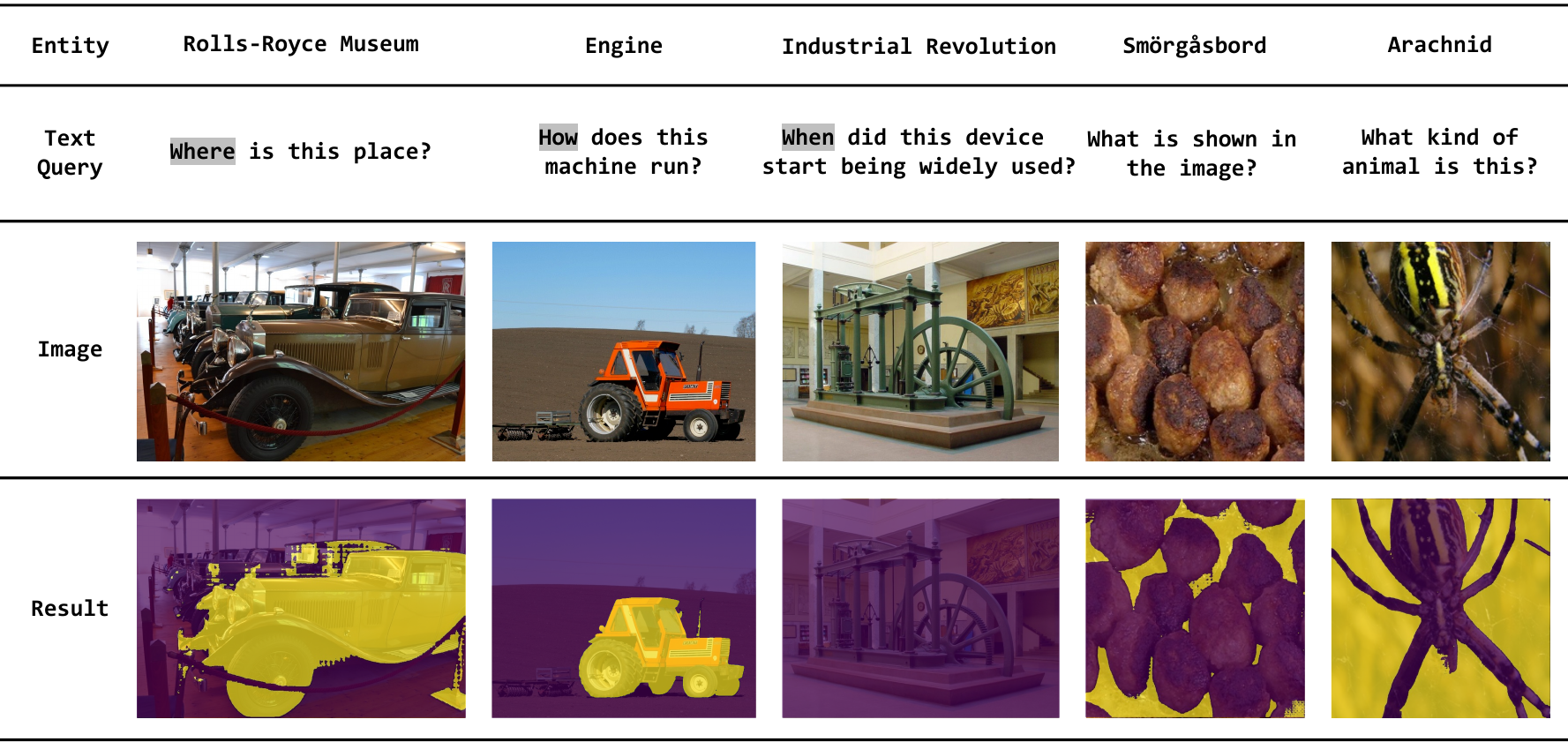}
\caption{Examples that triggered the filtering rules during the data annotation process.}

\label{fig:filter}
\end{figure*}

\section{Data Filtering}
\label{append:data_filtering}
The evaluation results in \cref{tab:anno_evaluation} demonstrate the effectiveness of our heuristic filtering rules based on model ensembles. This section provides further qualitative analysis and discusses the technical details. We identified 4 primary issues. Figure \ref{fig:filter} lists some cases corresponding to the above question. 
\begin{itemize}
    \item incomplete depiction of entities in images.
    \item references to non-visual entities.
    \item foreground-background confusion in dense object scenes.
    \item error propagation in the segmentation pipeline.
\end{itemize}

The first case involves an error of incomplete depiction of entities in images. The image shows a partial view of the entity \texttt{Rolls-Royce Museum}. This issue primarily occurs with entities of the `location' and `building' types. To address this, we set the pixel mask for such cases to cover the entire image.

The second and third cases fall under references to non-visual entities. The main issue involves either non-visual entities, such as \texttt{Industrial Revolution}, or those not visible in the image, such as \texttt{Engine}. To address this, we filter these errors based on the entity type and specific interrogative words in the text query. Specifically, we exclude entities of types such as time, location, method, event, game, and technology, as well as queries containing interrogative words like ``when," ``how," and ``why." As a result, we exclude 124,896 annotations in Entity Split, 7,920 in Query Split, and 176 in Human Set.

The fourth and fifth examples both involve foreground-background confusion but for different reasons. the fourth example is a typical dense object scene, while the fifth is due to error propagation. We apply different correction methods for these two types of errors. 

For dense objects, we first perform morphological transformations, including erosion and dilation, on the segmented masks. We then calculate their connected regions, and if the number exceeds the threshold, we classify it as a dense object scene. In such cases, we combine predictions from different models and use the confidence scores of these predictions to distinguish between foreground and background. 

For error propagation, the fifth example shows that foreground-background confusion arises because Grounding DINO \cite{liuGroundingDINOMarrying2023}predicts a bounding box that encompasses multiple objects. Consequently, this causes an error in the subsequent SAM \cite{kirillovSegmentAnything2023} step, which lacks a text prompt. To correct those cases, we use the annotation results from the end-to-end SEEM model \cite{zouSegmentEverythingEverywhere2023}.

\section{Additional Dataset Statistics}
\label{append:data_statistics}
Table \ref{tab:manual} presents the statistical information of the sample set used for the manual evaluation of annotation quality. The data samples are sourced from the Entity Split, Query Split, and Wiki Split. The sampling process involves two steps. First, we randomly select one sample from the annotated samples for each entity. Second, we sample based on the number of entities corresponding to each split from different dataset splits. For the Wiki Split, we randomly sample 200 instances. 
\begin{table}[ht]
    \centering
    \begin{tabular}{lrrr}
        \toprule
        \textbf{Split} & \textbf{Entity} & \textbf{Query} & \textbf{Wiki} \\ \midrule
        Case Num & 1400 & 400 & 200 \\
        Entity Num& 1400 & 400 & 200 \\ \bottomrule
    \end{tabular}
    \caption{Statistics for manual evaluation set.}
    \label{tab:manual}
\end{table}


Table \ref{tab:datasource} shows the detailed statistics of the Mask\textsc{Oven}-Wiki from 14 source datasets. Note that VQA v2 \cite{8100153} and OK-VQA \cite{8953725} are combined because their images are both sourced from COCO \cite{cocodataset}. These datasets primarily involve Visual Question Answering (VQA) and image retrieval tasks, and this distinction is also reflected in \cref{tab:datasource}.

\begin{figure*}
    \centering
    \subfigure[]{
        \includegraphics[width=0.46\columnwidth]{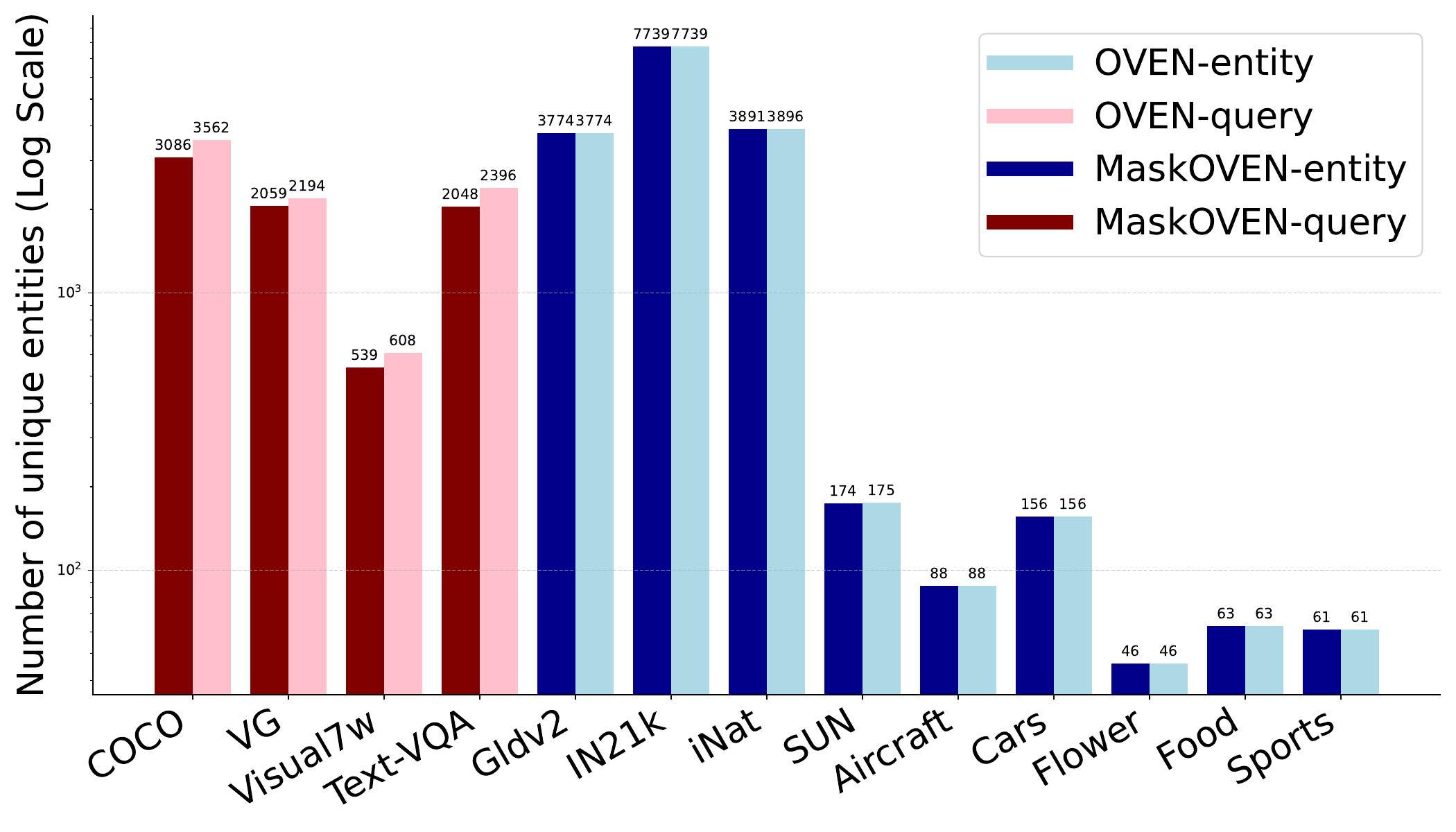}
        \label{fig:datasource1}
        \hspace{3mm}
    }
    \subfigure[]{
        \includegraphics[width=0.46\columnwidth]{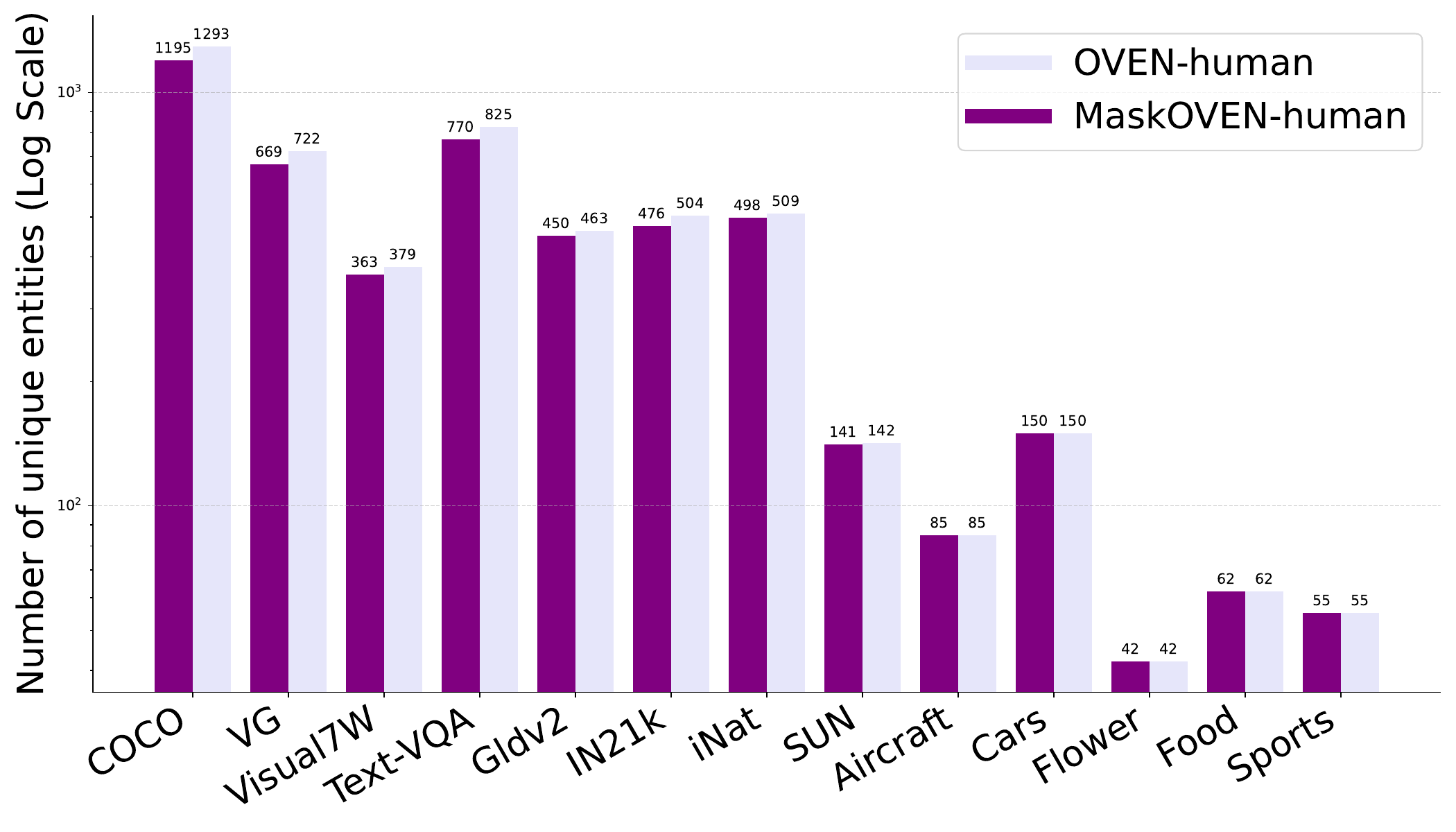}
        \label{fig:datasource2}
        \hspace{3mm}
    }
    \caption{Detailed statistics of unique entities grouped by source dataset on entity split (top red), query split (top blue), and human set (down purple). We compare them to the original statistics of \textsc{Oven}-Wiki.}
    \label{fig:datasource}
\end{figure*}

Figure \ref{fig:datasource} gives a comparison between Mask\textsc{Oven}-Wiki and \textsc{Oven}-Wiki \cite{huOpendomainVisualEntity2023} on the number of unique entities from different source datasets. This figure shows that the filtered entities are mainly distributed in query splits, and the source datasets of query splits mainly involve VQA tasks.
\begin{table*}
    \centering
    \begin{tabular}{lrrr}
        \toprule
        \textbf{Source} & \textbf{Entity Split} & \textbf{Query Split} & \textbf{Human Split} \\ \midrule
        IN21k \cite{deng2009imagenet} & 4,650,030 & 0 & 1,327 \\
        iNat \cite{Horn_2018_CVPR} & 276,130 & 0 & 1,591 \\
        Cars \cite{6755945}  & 12,481 & 0 & 1,605 \\
        SUN \cite{5539970}   & 21,646 & 0 & 1,275 \\
        Food \cite{food101}  & 29,193 & 0 & 1,513 \\
        Sports \cite{sport100} & 4,764 & 0 & 1,320 \\
        Aircraft \cite{maji2013finegrainedvisualclassificationaircraft} & 10,753 & 0 & 1,419 \\
        Flower \cite{4756141} & 2,711 & 0 & 1,405 \\
        Gldv2 \cite{weyand2020GLDv2} & 173,639 & 0 & 1,342 \\
        COCO \cite{cocodataset}* & 0 & 11,518 & 3,856 \\
        Visual7W \cite{7780909}* & 0 & 5,362 & 2,420 \\
        VG \cite{vg}* & 0 & 20,872 & 2,254 \\
        Text-VQA \cite{8953586}* & 0 & 3,165 & 1,830 \\ \bottomrule
    \end{tabular}
    \caption{Statistics for the amount of annotations in Mask\textsc{Oven}-Wiki from each source dataset. Datasets marked with * contribute to the VQA task, while the others contribute to the image retrieval task.}
    \label{tab:datasource}
\end{table*}

\end{document}